\documentclass[10pt,twocolumn,letterpaper]{article}

\usepackage{iccv}
\usepackage{times}
\usepackage{epsfig}
\usepackage{graphicx}
\usepackage{amsmath}
\usepackage{amssymb}
\usepackage{caption}

\usepackage{color}
\usepackage{colortbl}
\usepackage[dvipsnames]{xcolor}
\usepackage{mathtools}
\usepackage{amsthm}
\usepackage{pifont}%
\usepackage{multirow}
\newcommand{\cmark}{\ding{51}}%
\newcommand{\xmark}{\ding{55}}%
\usepackage{algorithm}
\usepackage{algorithmic}

\newcommand{\NNnote}[1]{}
\newcommand{\bw}[1]{}
\newcommand{\se}[1]{}

\theoremstyle{plain}

\theoremstyle{definition}

\theoremstyle{remark}

\usepackage[pagebackref=true,breaklinks=true,letterpaper=true,colorlinks,bookmarks=false]{hyperref}

\usepackage[capitalize,noabbrev]{cleveref}
\iccvfinalcopy

\usepackage{inconsolata}

\hypersetup{
   urlcolor=blue,
   citecolor=ForestGreen,
}

\ificcvfinal\fi

\title{End-to-End Diffusion Latent Optimization Improves Classifier Guidance}
\author{
Bram Wallace\\
Salesforce Research\\
{\tt\small b.wallace@salesforce.com}
\and
Akash Gokul\\
Salesforce Research\\
{\tt\small agokul@salesforce.com}
\and
Stefano Ermon\\
Stanford University\\
{\tt\small ermon@cs.stanford.edu}
\and
Nikhil Naik\\
Salesforce Research\\
{\tt\small nnaik@salesforce.com}
}
\begin{document}

\twocolumn[{%
\renewcommand\twocolumn[1][]{#1}%
\maketitle
\begin{center}
    \centering
    \captionsetup{type=figure}
    \includegraphics[width=0.9\textwidth]{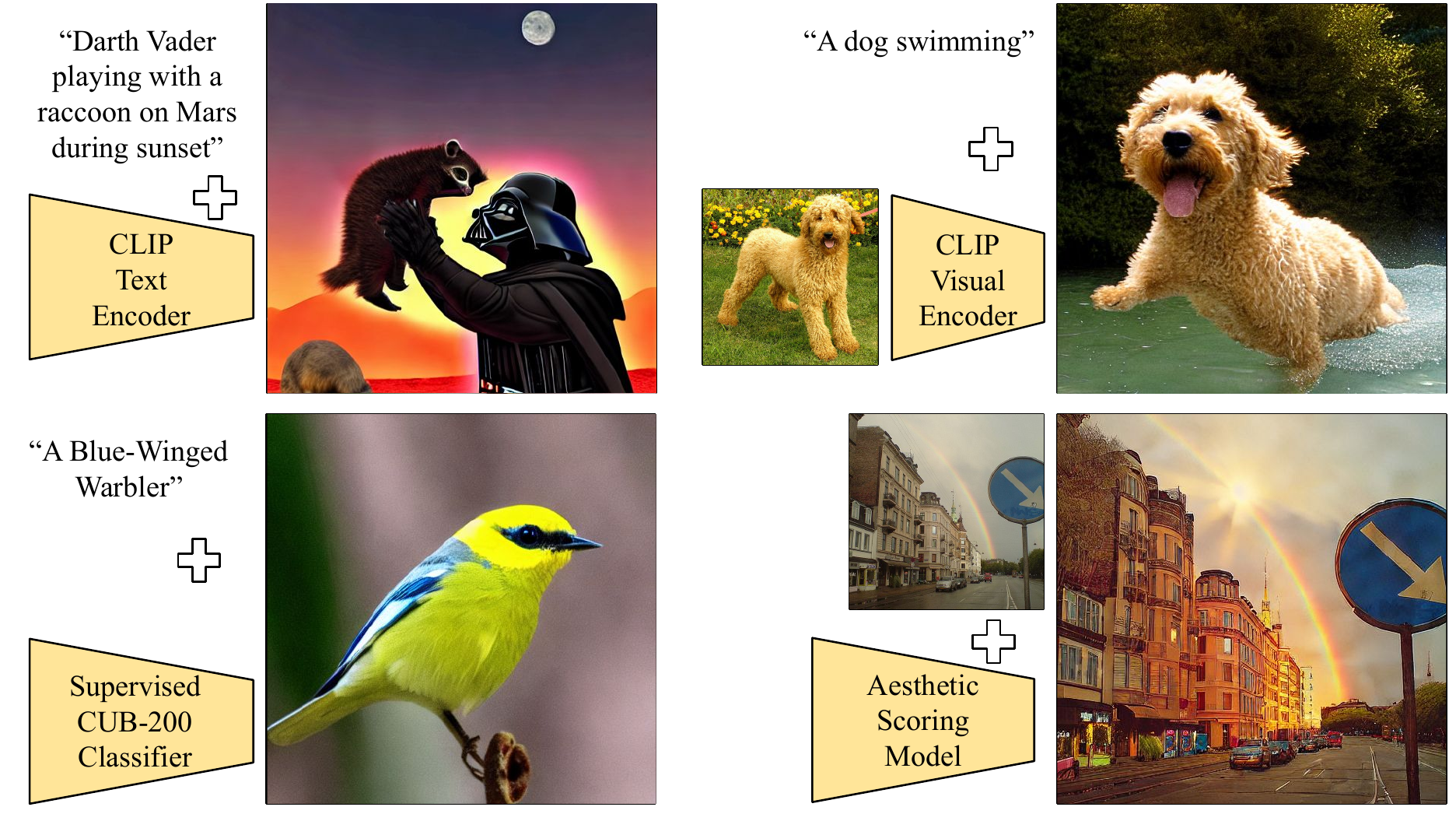}
    \captionof{figure}{We propose DOODL - a process that directly optimizes diffusion latents w.r.t. a model-based loss on the final generation. Our method improves on vanilla classifier guidance in all tested settings and we demonstrate capabilities novel to this class of methods such as vocabulary expansion, entity personalization, and perceived aesthetic value improvement.
    }
    \label{fig:splash}
\end{center}%
}]
\ificcvfinal\thispagestyle{empty}\fi

\begin{abstract}

Classifier guidance---using the gradients of an image classifier to steer the generations of a diffusion model---has the potential to dramatically expand the creative control over image generation and editing.
However, currently classifier guidance requires either training new noise-aware models to obtain accurate gradients or using a one-step denoising approximation of the final generation, which leads to misaligned gradients and sub-optimal control.%
We highlight this approximation's shortcomings and propose a novel guidance method:  Direct Optimization of Diffusion Latents (DOODL), which  enables plug-and-play guidance by optimizing diffusion latents w.r.t. the gradients of a pre-trained classifier on the true generated pixels, using an invertible diffusion process to achieve memory-efficient backpropagation. 
Showcasing the potential of more precise guidance, DOODL outperforms one-step classifier guidance on computational and human evaluation metrics across different forms of guidance: using CLIP guidance to improve generations of complex prompts from DrawBench, using fine-grained visual classifiers to expand the vocabulary of Stable Diffusion, enabling image-conditioned generation with a CLIP visual encoder, and improving image aesthetics using an aesthetic scoring network.
Code at \url{https://github.com/salesforce/DOODL}.
\end{abstract}
\vspace{-4mm}

\section{Introduction}

Text-conditioned denoising diffusion models (DDMs), such as Latent/Stable Diffusion, Imagen and numerous others, have been widely adopted for synthesizing realistic images given an input text prompt~\cite{ldm,stablediffusion2,imagen,dalle2,ediffi}.
The \textit{conditioning} setup requires DDMs to be trained using paired data of images and the conditioning modality---image-caption pairs in the case of text conditioning. Once trained, the DDM can be steered to generate images using the conditioning modality. However, the \textit{conditioning} paradigm constricts the image generation capabilities of a trained DDM model; a text-conditioned diffusion model cannot readily utilize other modalities such as depth maps or image classifiers or audio as conditioning signals. While DDMs conditioned on multiple modalities have been trained~\cite{imagevariations,stablediffusion2} a ``plug-and-play'' approach that allows for a pretrained DDM to be guided by any external function that determines whether some generation criterion is satisfied is desirable.

\begin{figure}
    \centering
    \includegraphics[width=\linewidth]{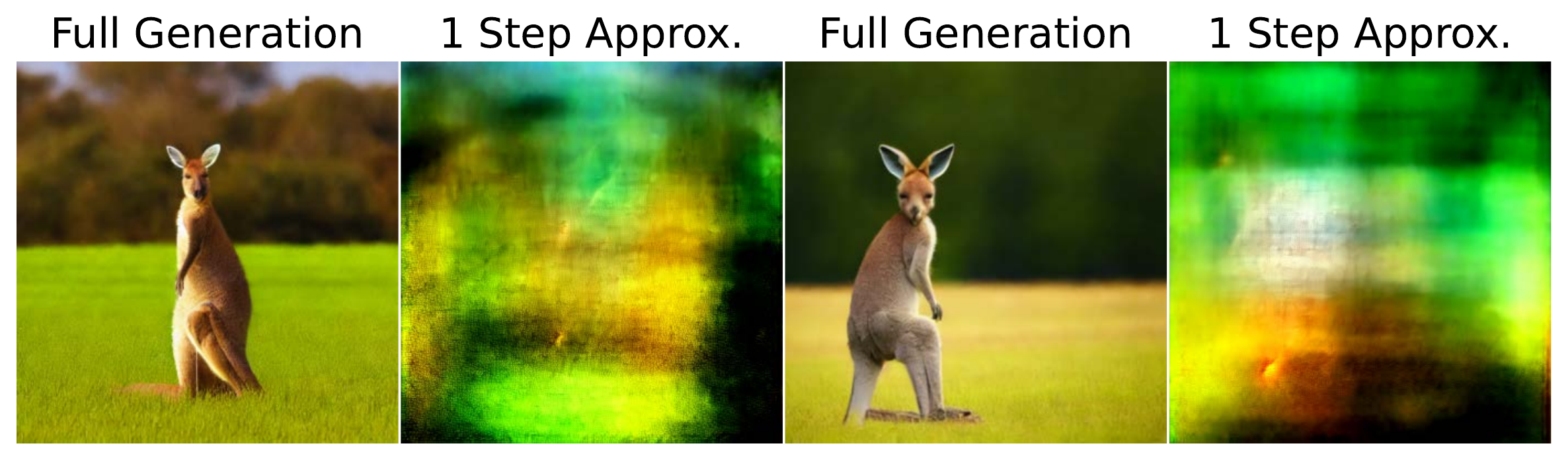}
    \caption{Stable Diffusion with prompt \textit{``A kangaroo in a field''} using 50 DDIM steps. vs. a single step.
    The one-step approximation is inaccurate for high noise levels.}
    \label{fig:misalignment}
\end{figure}

In principle, \textit{classifier guidance}~\cite{song2020score,dhariwal2021diffusion} enables this capability in DDMs. Classifier guidance, named so since it was first demonstrated using pretrained image classification models, combines the score estimate of the diffusion model with the gradient of the image classifier to steer the generation process to produce images that correspond to a particular class. Any differentiable loss function can be used for classifier guidance. In addition to class-conditional generation,  it has been shown to improve compositionality using cross-modal guidance~\cite{upainting}. There are two existing ways of incorporating classifier guidance. In the first approach, we train a {noise-aware} classifier that can be used to compute an accurate gradient w.r.t. an intermediate generation step~\cite{song2020score,dhariwal2021diffusion}. This approach requires re-training the classifier model, which can be computationally expensive/infeasible due to lack of access to training data. In the second approach,  at a time step $t$, we denoise the image with a single application of the DDM and compute the gradient using this approximately denoised image~\cite{upainting}. 
This one-step approximation is necessitated by the prohibitive memory requirements of computing a gradient w.r.t. the latents through the entire diffusion process containing many steps.  
Since this approach only obtains gradients using an one-step denoising approximation of the final generation, the approximately denoised images are often misaligned with the final generations which classifier guidance is aiming to modify (\cref{fig:misalignment}), leading to sub-optimal guidance signal.

To enable flexible and exact model guidance, without noise-aware classifiers or approximations, we propose Direct Optimization Of Diffusion Latents (DOODL). DOODL optimizes the initial diffusion noise vectors w.r.t. a model-based loss on images generated from the full-chain diffusion process. We leverage EDICT, a recently developed drop-in \textit{discretely invertible} diffusion algorithm~\cite{edict}, which admits backpropagation with constant  memory cost w.r.t the number of diffusion steps,  to compute classifier gradients on the pixels of the final generation w.r.t. the original noise vectors.
This enables {efficient} iterative optimization of diffusion latents w.r.t. any differentiable loss on the image pixels and accurate calculation of  gradients for classifier guidance.

We demonstrate the efficacy of DOODL through a diverse set of guidance signals (\cref{fig:splash}) using quantitative and human evaluation studies. First, we show that CLIP classifier guidance using DOODL improves generation of images guided by text prompts from the DrawBench~\cite{imagen} dataset, which test compositionality and the ability to guide using unusual captions. Second, we show the ability to expand the vocabulary of a pretrained Stable Diffusion model using fine-grained visual classifiers; a capability that one-step classifier guidance does not have. Third, we demonstrate that DOODL can be used for personalized entity generation (e.g. ``A dog in sunglasses''), \textit{with zero retraining of any new network}---a first to our knowledge. Finally, we utilize DOODL to perform a novel task; increasing the perceived aesthetic quality of generated/real images. We hope that DOODL can enable and inspire diverse plug-and-play capabilities for pretrained diffusion models. 

\section{Related Work}

\subsection{Text-to-Image Diffusion Models}

Text-to-image diffusion models~\cite{ddpm, song2020score, song2021maximum} such as GLIDE~\cite{glide}, DALLE-2~\cite{dalle2}, Imagen~\cite{imagen,imagenvideo}, Latent Diffusion~\cite{ldm,stablediffusion2}, and eDiffi~\cite{ediffi} have recently emerged at the forefront of image generation, based on methods in non-equilibrium thermodynamics~\cite{thermo}. 
Classifier guidance~\cite{song2020score,dhariwal2021diffusion}, and its adoptions~\cite{glide,upainting}, use the gradients of pretrained classifier models to guide such generations. 
Instead of sequential denoising, ~\cite{song2020score} traverses at a fixed noise level before each denoising step.
Concurrent work~\cite{universalguidance} modifies classifier guidance to refine the gradient prediction at each noise level before continuing.
\cref{tab:rel_work_comp} shows the requirements for learned methods such as ControlNet~\cite{controlnet} vs. these guidance-based methods. The former require \textit{data} and \textit{training}, while the latter substitute this need for \textit{pretrained recognition models}.
Directly optimizing latent variables of other generative models such as GANs or VAEs~\cite{vqgan,clip,vaelatentop} w.r.t. a pixel-based loss has been shown to generate in-distribution images that fit targeted criteria.
To the best of our knowledge, these techniques have not been applied to diffusion models before DOODL. %

At the intersection of diffusion models and invertible neural networks is a recently proposed approach called EDICT~\cite{edict} which algorithmically reformulates the denoising diffusion process to be {invertible}. This prior work focuses solely on the applications to image editing and does not consider properties of Invertible Neural Networks (INNs) or similar processes.
Methods such as DDIM are theoretically invertible in the limit of discretization, but this limit cannot be achieved in practice~\cite{edict}.

\begin{table}[]
    \centering
\resizebox{\linewidth}{!}{%
    \begin{tabular} {>{\centering\arraybackslash}m{4cm} | >{\centering\arraybackslash}m{1cm} >{\centering\arraybackslash}m{1cm} >{\centering\arraybackslash}m{2cm}}
        & \multicolumn{3}{c}{\centering \textbf{Requirements}} \\
        {Method Type}  & Training & {Data} & Pretrained ``Classifier'' \\
        \hline
         \textit{Learning-Based}  & \multirow{2}{*}{\cmark} & \multirow{2}{*}{\cmark} & \multirow{2}{*}{\xmark} \\
         ~\cite{textualinversion,dreambooth,controlnet} \\
         \hline
         \textit{Guidance-Based} & \multirow{2}{*}{\xmark} & \multirow{2}{*}{\xmark} & \multirow{2}{*}{\cmark} \\
         (\textbf{DOODL}, Clf. Guidance) 
    \end{tabular}
    }
    \caption{
    \textit{Learning-Based} methods require data and training while \textit{Guidance-Based} methods require pretrained recognition networks (trained on non-noisy data in our setting).
    }
    \label{tab:rel_work_comp}
\end{table}

\subsection{Invertible Neural Networks (INNs)}\label{sec:related_inv}

While neural networks tend to be non-dimensionality-preserving functions, there has been prior work on constructing reversible architectures.
A predominant class of such INNs are normalizing flow models~\cite{nice,realnvp,glow}.
A modified version of the ``coupling layers'' in normalizing flow architectures are incorporated into the EDICT~\cite{edict} algorithm employed in this work. 
~\cite{iresnet} proposes an architecture guaranteed to be invertible via well-conditioned inverse problems instead of a closed-form solution.
The memory savings of such architectures has been used in long-sequence recurrent neural networks~\cite{mackay2018reversible} and to study inverse problems~\cite{inverseproblems}.

\section{Background}
\subsection{Invertible Neural Networks w.r.t Memory}\label{sec:inns}

When using gradient descent for optimizing neural networks, with network parameters $\Xi = \{\xi_p \}^{p=P}_{p=1}$, network input $x$,  network output $y=f(x)$, and loss function $c$, the derivative $\frac{dc(y)}{d\xi}$ is calculated and gradient descent performed to minimize $\mathbb{E}_{Data}c(y)=\mathbb{E}_{Data}c(f(x))$. Here $f$ is implicitly conditioned on $\Xi$.
Consider $f$ as the composition of $n$ functions (layers) $f^{n} \circ f^{n-1} \circ ... \circ f^{1}$.
To optimize $\xi$, a parameter of the $i^{th}$ layer $f^i$ the derivative $\frac{dc(y)}{d\xi}$ is calculated.
Denote $f^{k} \circ f^{k-1} \circ ... \circ f^{j}=F^{k}_{j}$.
Since $y= F^{n}_{1}(x)$ the derivative w.r.t. $\xi$ can be calculated using the chain rule: 
\begin{align}\label{eq:chain}
    & \frac{dc(y)}{d\xi} = \frac{dc(F^{n}_{1}(x))}{d\xi} \\
    & = \frac{dc(F^{n}_{1}(x))}{dF^{n}_{1}(x)} \cdot \frac{dF^{n}_{1}(x)}{dF^{n-1}_{1}(x)}\cdot \cdot \cdot \frac{dF^{i}_{1}(x)}{dF^{i-1}_{1}(x)} \cdot \frac{dF^{i-1}_{1}(x)}{dx}
\end{align}
In the general case, calculating $\frac{dc(y)}{d\xi}$ requires storing all intermediate activations, a key bottleneck to backpropagation.
Network sharding~\cite{sharding} across processors reduces the per-processor hardware memory requirement but the total remains the same.
Gradient checkpointing~\cite{checkpointing1,checkpointing2} decreases memory cost, but increases runtime linearly w.r.t. saved memory.
INNs can recover intermediate states/inputs from the output, reducing memory costs by avoiding activation caching.
If every $f^{j}$ is invertible in \cref{eq:chain}, the denominator terms can be reconstructed during the backwards pass; such methods have been leveraged to train large INNs much faster than non-invertible equivalents(\cref{sec:related_inv}).

\subsection{Denoising Diffusion Models (DDMs)}
Image DDMS are trained to predict the noise $\epsilon$ added to an image $x$ ~\cite{ddim,ddpm,dhariwal2021diffusion,classifierfree,song2019generative}. 
Noise levels are discretized into a set $\mathcal{T}=\{0, 1, .. ,T\}$ that index a noising schedule $\{\alpha_t\}^{T}_{t=0}, \alpha_T=0, \alpha_0=1$. $t\in\mathcal{T}$ are randomly sampled during training and paired with data $x^{(i)}$ (images or autoencoded representations) to generate noisy samples
\begin{equation}\label{eq:diffusion_objective}
    x^{(i)}_t = \sqrt{\alpha_t} x^{(i)} + \sqrt{1 - \alpha_t} \epsilon
\end{equation}
where $\epsilon \sim N(0, I)$.
DDMs conditioned on the timestep $t$ and auxiliary information (e,g, image caption) $C$ are trained to approximate the added noise $DDM(x^{(i)}_t, t, C) \approx \epsilon$. 
At generation time, an $x_T \sim N(0,1)$ is drawn and the DDM is iteratively applied to hallucinate a real image from the noise. %
Following the DDIM~\cite{ddim} sampling model, the final generation $x_0$ is equal to the composition of $S$ denoising functions: applications of $\Theta$ conditioned on $C$ and varying timesteps $t$.
Denoting $\Theta(x, t, C)$ as $\Theta_{(t,C)}(x)$
\begin{align}\label{eq:ddim}
x_0 = [ \Theta_{(0, C)} \circ \Theta_{(1, C)} \circ ... \circ \Theta_{(T, C)} ] (x_T)
\end{align}

\subsubsection{Classifier Guidance}\label{sec:clf_guidance}
In addition to $C$, other guidance signals can steer generations to objectives.
The foremost example, ``classifier guidance'' incorporates  gradients of a loss ($c_{clf}$, from a classifier network $\Phi$) on estimated pixels into the noise prediction~\cite{song2020score,dhariwal2021diffusion}.
From a theoretical perspective, this typically is the gradient of the log-conditional probability $\nabla \log p_{\Phi}(y|x_t)$
There are two primary ways of incorporating classifier guidance:

\noindent \textbf{1:} A \textit{noise-aware} classifier is trained for direct use on intermediate (noisy) $x_t$, with $\nabla_{x_t} c_{clf}(x_t)$ incorporated into the denoising prediction~\cite{glide}.
Training noise-aware models is effective but often infeasible due to computational expense and data availability (e.g. medical images). This results in there being very few publicly available noise-aware models.

\noindent \textbf{2:} $x_0$ is approximated with a single application of $\Theta_{(t, C)}$ as in the training objective~\cite{upainting}. Here,  the incorporated gradient  is $\nabla_{x_t} c_{clf}(x^*_0)$, where $x^*_0$ is a 1-step approximate solve by substituting $\epsilon$ for $\Theta_{(t,C)}$ in \cref{eq:diffusion_objective}.
While a standard (noise-unaware) model can be used, the gradients are calculated w.r.t. an \textit{approximation} of $x_0$ (\cref{fig:misalignment}) and as such may be misaligned with the true derivative of $\frac{dc_{clf}(x_t)}{x_0}$.

\subsubsection{Exact Inversion of the Diffusion Process}\label{sec:edict}
Recently, EDICT~\cite{edict}, an exactly invertible  variant of the discrete (time-stepped) diffusion process was proposed. 
EDICT operates on a latent pair $(x_t,y_t)$ instead of a single variable. %
Initially $x_T=y_T \sim N(0,I)$, followed by iteratively denoising using the reverse diffusion process:
\begin{align}
\begin{split}\label{eq:edict_gen}
    x^{inter}_t &=  a_t \cdot x_{t} + b_t \cdot  \Theta_{(t,C)}(y_t) \\
    y^{inter}_t &=  a_t \cdot  y_{t} + b_t  \cdot  \Theta_{(t,C)}(x^{inter}_t) \\
    x_{t-1} &= p \cdot x^{inter}_t + (1-p) \cdot y^{inter}_t \\
    y_{t-1} &= p \cdot y^{inter}_t + (1-p) \cdot x_{t-1} \\
\end{split}
\end{align}
where ($a_t$, $b_t$) are time-dependent coefficients, and $p \in [0,1]$ is a mixing parameter to mitigate latent drift. Intuitively, this process first updates the $x$ and $y$ sequences based on the current state of the counterpart, and then invertibly ``averages'' them together.
The above equations admit linear solves to invert them, defining the inverse process:
\begin{align}
\begin{split}\label{eq:edict_noise}
    y^{inter}_{t+1} &= ({y_{t} - (1-p) \cdot x_{t}})/{p} \\
    x^{inter}_{t+1} &= ({x_{t} - (1-p) \cdot y^{inter}_{t+1}})/{p} \\
    y_{t+1} &= ({y^{inter}_{t+1} - b_{t+1}\cdot  \Theta_{(t+1,C)}(x^{inter}_{t+1})  })/{a_{t+1}} \\
    x_{t+1} &= ({x^{inter}_{t+1} - b_{t+1} \cdot \Theta_{(t+1,C)}(y_{t+1})})/{a_{t+1}} \\
\end{split}
\end{align}
We employ this construction in DOODL, and use \cref{eq:edict_noise} to encode images $x_0$ to latents $x_T$ in \cref{sec:aes}.

\begin{figure*} 
    \centering
    \includegraphics[width=.97\linewidth]{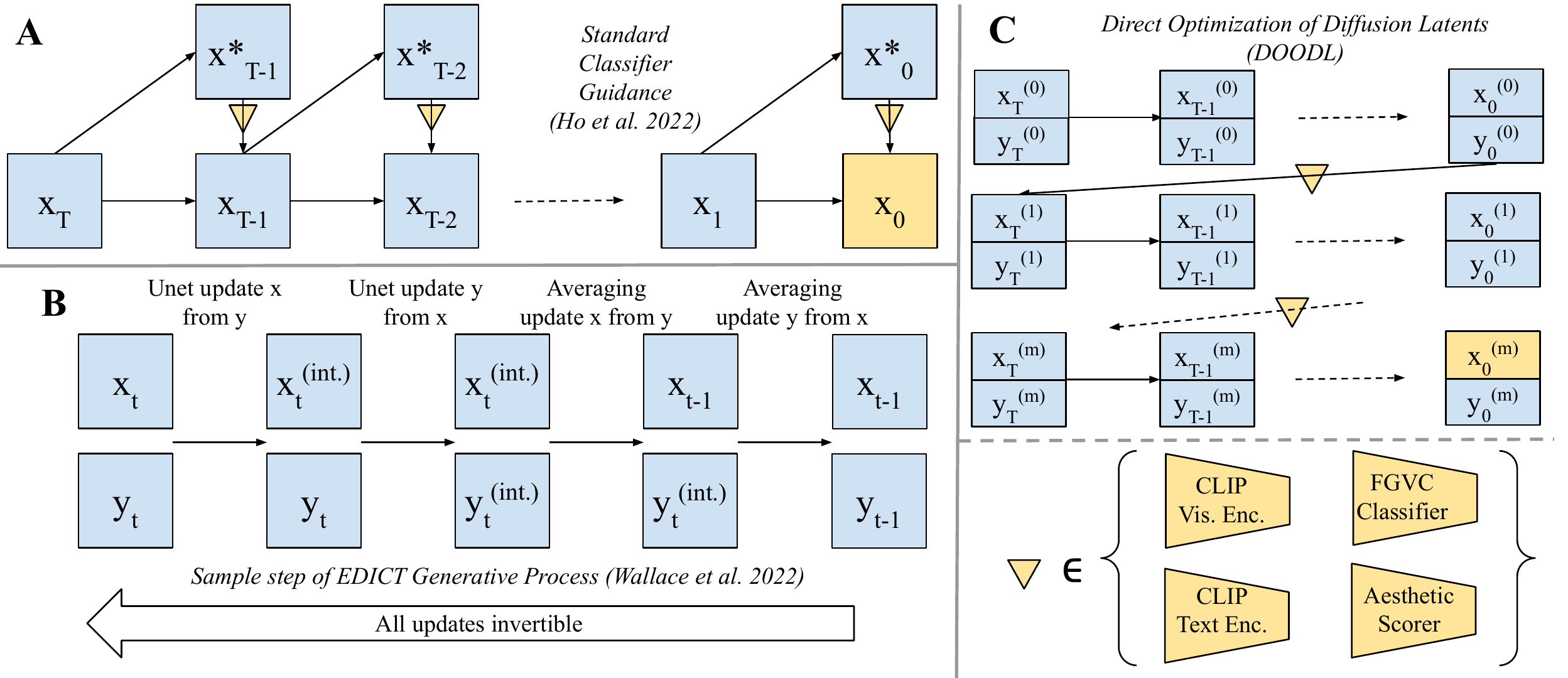}
    \caption{Method diagrams. \textbf{A} Standard classifier guidance: at each timestep, $t$, a one-step denoising approximation of $x_0$ is computed and the loss is calculated w.r.t the pixels of this generation. The gradient of this loss is incorporated into the subsequent diffusion step. \textbf{B} EDICT~\cite{edict}, an invertible variant of the diffusion process which admits backpropagation through the entire chain with no additional memory cost. \textbf{C} DOODL, our proposed method. We leverage EDICT and demonstrate that the gradients of model losses computed w.r.t. the final generation can be used to optimize the fully noised $x_T$ directly.
    { \large \textcolor{yellow}{$\blacktriangledown$} } indicates a gradient calculation from a differentiable model-based loss with networks employed in this work displayed.
    }
    \label{fig:diagram}
\end{figure*}

\section{Direct Optimization of Diffusion Latents}\label{sec:method}
We aim to develop a method that overcomes the shortcomings of classifier guidance discussed in \cref{sec:clf_guidance}. Concretely, a method that (1) does not require re-training/finetuning an existing pretrained classifier model, (2) computes gradients w.r.t. the {true} output instead of a one-step approximation, and (3) incorporates the guidance in a semantically meaningful way (as opposed to an adversarial-style perturbation). 
We emphasize the last point in particular. As shown in the literature on adversarial attacks~\cite{adversarial1,adversarial2,adversarial3},  gradients w.r.t. pixels can satisfy a classifier loss while not perceptually changing the content of an image. 
This is as opposed to techniques such as latent optimization in GANs, where the regularization provided by the decoder means that optimization happens in a space where perturbations typically result in perceptually meaningful changes that satisfy the desired objective.
In this work, we aim to directly optimize \textit{diffusion latents}, a first in the literature to our knowledge.

As shown in \cref{eq:ddim}, it is mathematically trivial to optimize $x_T$ for a desired outcome on $x_0$. There is a closed form expression for $\frac{dx_0}{x_T}$ as in \cref{eq:chain}.
However, due to activation caching, the na\"ive memory cost is linear in the number of DDIM sampling steps due to the $T$ applications of $\Theta$.
With a typical value of $S=50$, this memory cost nears a terabyte for state-of-the-art diffusion models which is impractical for most uses.
Gradient checkpointing trades memory for computational complexity, resulting in the computational complexity of each backwards pass increasing by a factor of $S$ if memory costs are held constant.

We draw inspiration from INNs (\cref{sec:inns}) to optimize $x_T$ w.r.t. criteria on $x_0$ in feasible runtime.
Using invertible $\Theta_{(i,C)}$ in \cref{eq:ddim}, intermediate states can be reconstructed during the backwards pass using only a constant number of applications of $\Theta$ w.r.t. $T$, circumventing the prohibitive memory cost without sacrificing runtime.

We turn to the recently developed EDICT~\cite{edict} as an invertible reverse diffusion process that admits constant-memory implementation of optimization of $x_t$. 
Given conditioning $C$, differentiable model-based cost function $c$ and a latent draw $x^{(0)}_T$, the EDICT generative process is performed ( 50 steps, $p=0.93$, StableDiffusion v1.4) yielding initial output $f(x^{(0)}_T) = x^{(0)}_0$, which is then used to calculate a loss $c(x^{(0)}_0)$ and corresponding gradient $\nabla_{x_t} c(f(x^{(0)}_T))$. This gradient can then be used to perform a step of gradient descent optimization on $x^{(0)}_T$. %

We modify vanilla gradient descent in several key ways to achieve realistic images that satisfy the guidance criteria.
After each optimization step, the EDICT ``fully noised'' latent pair $x^{(j)}_T$ and $y^{(j)}_T$ (from \cref{eq:edict_gen,eq:edict_noise}) are averaged together and renormalized to the original norm of the initial draw $x^{(0)}_T$.
The averaging prevents latent drift which degrades quality, as noted in~\cite{edict}. 
Normalizing to the original norm keeps the latent on the ``gaussian shell'' and in-distribution for our diffusion model. %

We also perform multi-crop data augmentation on the generated $(x_0, y_0)$, sampling 16 crops per image (details in Supplementary).
Momentum with $\eta=0.9$ is employed. We do not find Nesterov momentum to be useful.
Finally, to increase stability and realism of outputted images, we perform element-wise clipping of $g$ at magnitude $10^{-3}$ and perturb $x_T$ by $\mathcal{N}(0, 10^{-4} \cdot I)$ at each update.
Our algorithm is formally laid out in \cref{alg:doodl}.
\cref{fig:diagram} provides an overview of our method alongside classifier guidance.

\section{Applications and Results}

Previous work has used dataset- or webly-supervised classifiers such as CLIP~\cite{clip, glide}. We consider these as well as previously unexplored settings. 
The optimization learning rate, $\lambda$, is a key hyperparameter and we note its values throughout. %
The focus of our work is on how the gradients from \textit{non-noise-aware} classifiers can be better incorporated into the diffusion process. As such, we do not consider noise-aware classifier methods~\cite{glide}.
For larger images and other example results, please refer to the Supplementary.

\begin{algorithm}
   \caption{DOODL}
   \label{alg:doodl}
\begin{algorithmic}
   \STATE {\bfseries Input:} $\lambda$ (Learning rate), $C$ (Model conditioning), $c$ (Cost function), $T$ (\# Diffusion steps), $m$ (\# Optimization steps), $\Theta$ (Diffusion Model)
   \STATE Initialize $x^{(0)}_T = \mathcal{N}(0, I)$.
   \STATE $g_{-1}=0$
   \FOR{$i=0$ {\bfseries to} $m$} 
   \STATE $x_T, y_T = x^{(i)}_T$
   \FOR{$t=T-1$ {\bfseries to} $0$} %
        \STATE $x^{inter}_t =  a_t \cdot x_{t} + b_t \cdot  \Theta_{(t,C)}(y_t) $ \\
        \STATE $y^{inter}_t =  a_t \cdot  y_{t} + b_t  \cdot  \Theta_{(t,C)}(x^{inter}_t) $\\
        \STATE $x_{t-1} = p \cdot x^{inter}_t + (1-p) \cdot y^{inter}_t $\\
        \STATE $y_{t-1} = p \cdot y^{inter}_t + (1-p) \cdot x_{t-1} $\\
   \ENDFOR
   \STATE $L = 0.5 \cdot (c(MultiCrop(x_0)) + c(MultiCrop(y_0)))$
   \STATE $g_i = - \lambda \cdot \nabla_{x_T} L$ \textbf{Linear in $T$ b/c of EDICT}
   \STATE $g_i = Clip(g_i, -10^{-3}, 10^{-3})$
    \STATE $g_i = 0.9 \cdot g_{i-1} + g_i$
   \STATE $x^{(i+1)}_T = x_T + g_i + \mathcal{N}(0, 10^{-4} \cdot I)$
   \ENDFOR
   \STATE {\bfseries yield} $x^{(m)}_T$ 
\end{algorithmic}
\end{algorithm}

\subsection{Reinforcing Text Guidance}\label{sec:text_guidance_method}

Classifier guidance using a CLIP text-image similarity loss has been used to reinforce the text conditioning $C$ provided as input to a text-to-image diffusion model~\cite{glide}. 
Previous work~\cite{upainting} shows that classifier guidance with gradients calculated on 1-step approximations of $x_0$ enhance the ability to generate complex prompts.
This setting serves as an initial validation of our proposed method.

We employ a spherical distance loss from the literature:
\begin{align}\label{eq:spherical_dist}
    d(x,y) = 2\lambda \sqrt{sin^{-1}(\frac{||x-y||}{2})}
\end{align}
computing a distance between the embedding of $C$ by the CLIP text encoder, $CLIP_{Text}(C)$, and the embedding of $x^{(i)}_0$ by the image encoder, $CLIP_{Image}(x^{(i)}_0)$.
$\lambda \in [10,100]$ is effective for the baseline one-step guidance, we use $\lambda=30$. 
For DOODL, we find $\lambda=0.1$  to be preferable. We attribute the lower value to the gradient magnitudes increasing when computed through repeated applications of the network, as well as the incorporation of momentum. 

\begin{figure}
    \centering
    \includegraphics[width=0.95\linewidth]{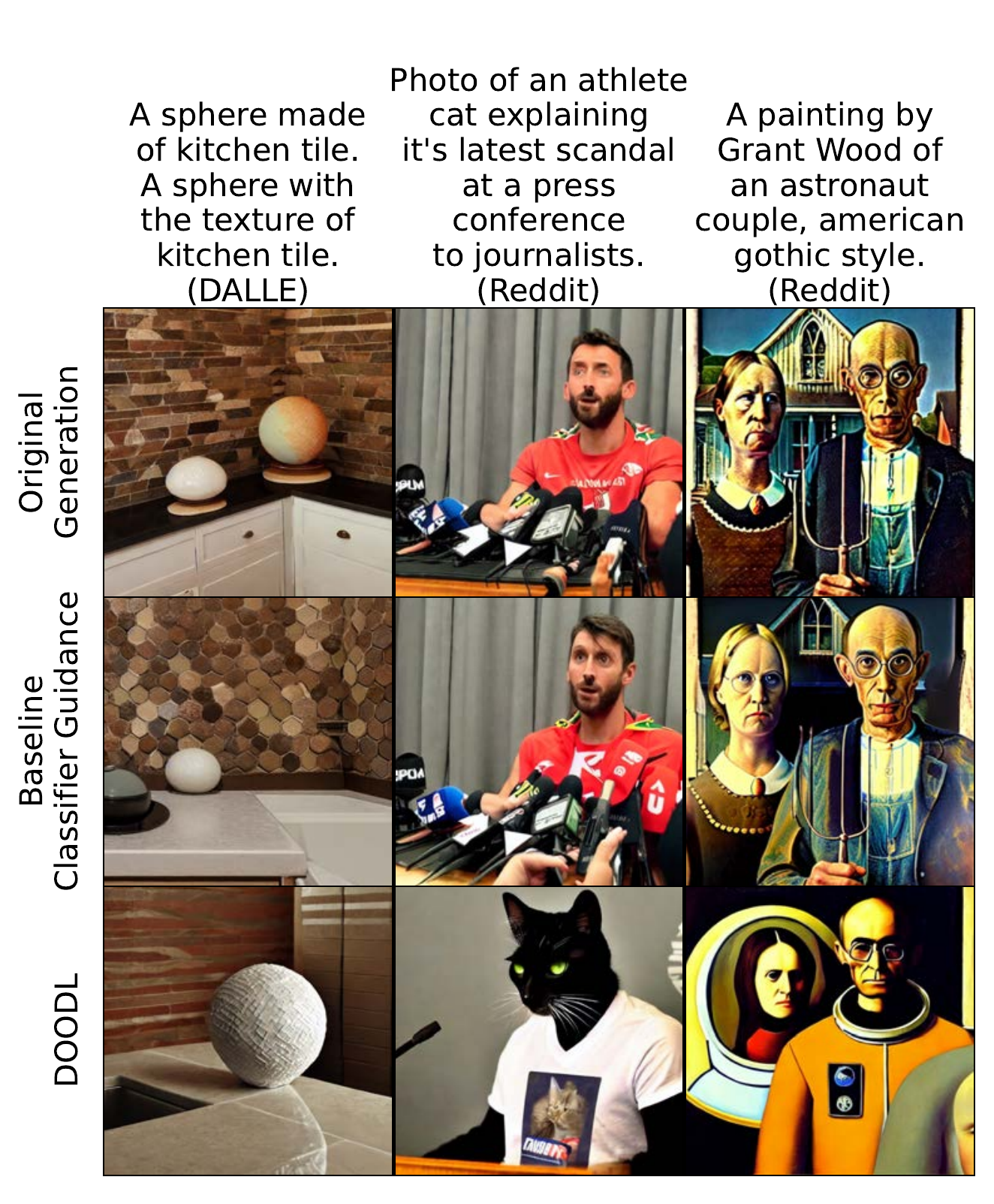}
    \caption{
    Generations using  DrawBench prompts\cite{imagen} for DOODL and baselines from the same seed.
    Note the prompting style of the \textit{DALLE} category uses multiple sentences to reinforce a single concept.
    }
    \label{fig:DrawBench_qualitative}
\end{figure}

\begin{figure}
    \centering
    \includegraphics[width=\linewidth]{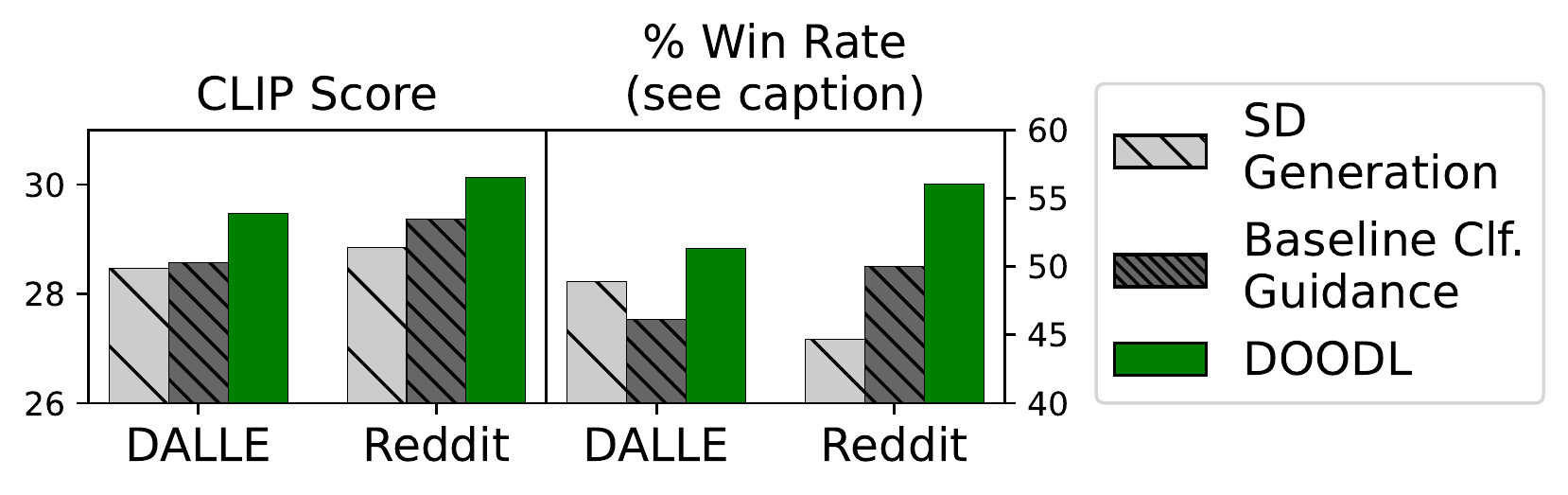}
    \caption{DrawBench results. \textbf{Left:} CLIP Scores. \textbf{Right:} Human evaluation.
    5 labelers are given a pairwise comparison of DOODL to another method for the same prompt-seed inputs and asked which represents the prompt better, with options for equal success or failure. We display the fraction of time for each method that it is classified by the majority of labelers as ``Better" or ``Both Achieve". Value for DOODL averaged across two comparison runs. }
    \label{fig:DrawBench_quant}
\end{figure}

We test on the DrawBench benchmark~\cite{imagen}, a collection containing 11 categories of text prompts designed to probe different aspects of text-to-image generation.
In particular, we evaluate on the  \textit{DALLE} and \textit{Reddit} categories of DrawBench, which focus on compositionality and highly implausible scenes, being especially suited for testing fidelity to conditioning text. 
Image-text alignment is measured both automatically (CLIP score) and with human evaluation in \cref{fig:DrawBench_quant}.
Experiments are performed across 9 seeds for all method-prompt pairs.
We use a LAION-trained CLIP model~\cite{laionclip} for classifier guidance and an OpenAI-trained CLIP model~\cite{clip} for CLIP score evaluation. 
While there is an overlap in the training sets of these models; we find them sufficiently independent to provide automated validation.
 Generated samples are shown in~\cref{fig:DrawBench_qualitative}.

DOODL relatively improves on baseline classifier guidance's CLIP score by 3.1\% and 2.6\% on \textit{DALLE} and \textit{Reddit} respectively.
For human study, we define prompt alignment success rate (PASR) as the percent of generations rated as humans to be aligned with the prompt in head-to-head comparisons (see \cref{fig:DrawBench_quant} caption).
The PASR of DOODL is highest, notably increasing on the complex unusual prompts of \textit{Reddit} by 11.3\% as compared to vanilla Stable Diffusion, with a 6\% increase over one-step classifier guidance.  

\begin{figure}
    \centering
    \includegraphics[width=\linewidth]{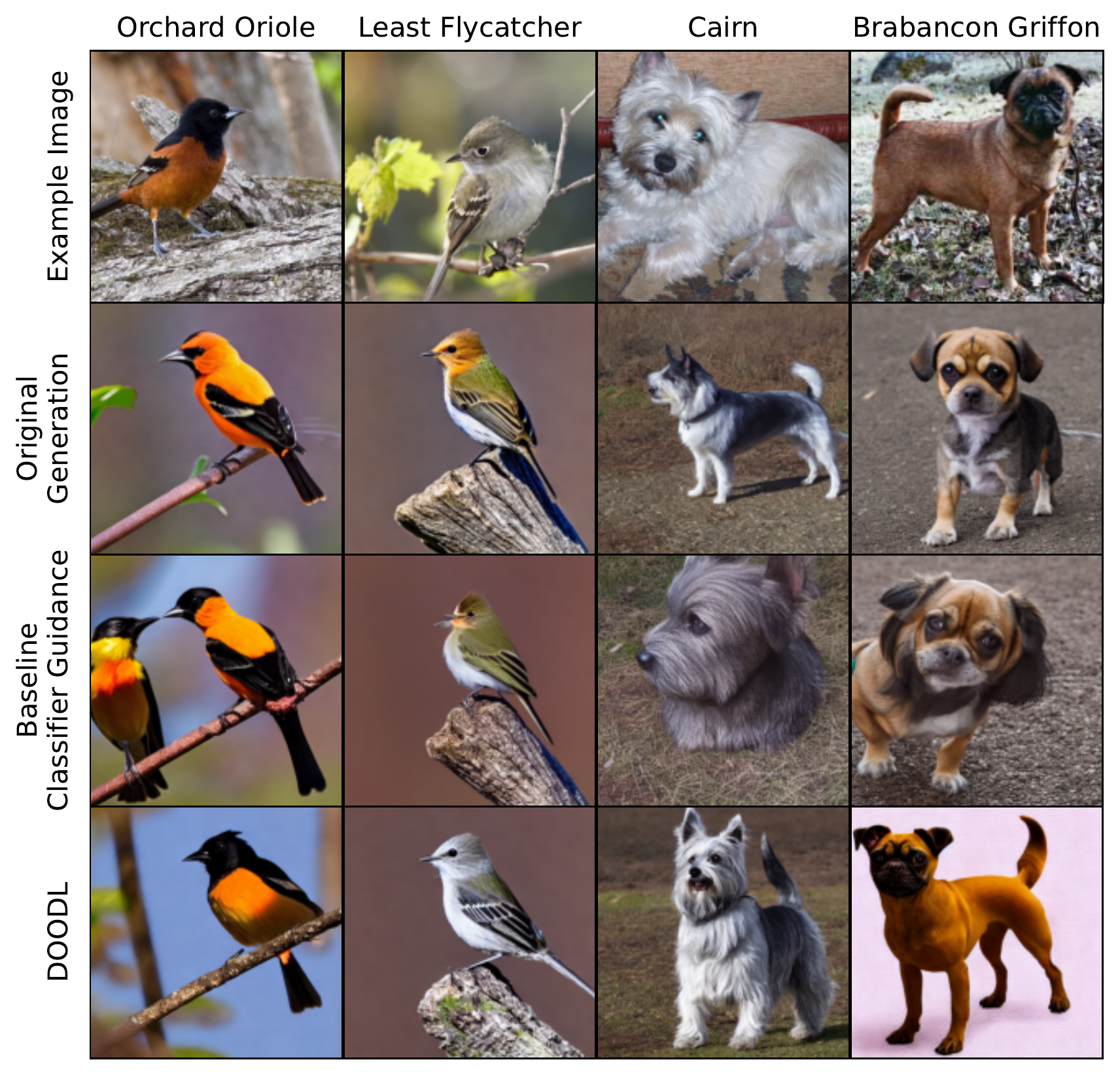}
    \caption{
    Qualitative generations for fine-grained classes with rare vocabulary. \textit{Example Image} is an exemplar of the class from the CUB~\cite{cub}/Dogs~\cite{dogs} dataset.
    Though the optimization process of DOODL starts from pixels not matching the target class (\textit{Original Generation}) due to the rare vocabulary failure of StableDiffusion, it is able to produce an image of the targeted species using the gradients of a supervised classifier trained solely on the specialized dataset.
    }
    \label{fig:fgvc_qualitative}
\end{figure}

\begin{table}[]
    \centering
    \begin{tabular}{c | c c c}
     \multicolumn{4}{c}{\centering Per-Class \% FID Change (Lower Better)} \\
     \hline
         Dataset & CUB & Dogs & Aircraft  \\
         \hline
         DOODL &  \textbf{-3.2\%} & \textbf{-5.3\%} & \textbf{-0.4\%} \\
         Baseline & +6.1\% & +0.27\% & +13.7\% \\  
    \end{tabular}
    \caption{Quantitative results for rare vocabulary generation. %
    Per-class FID is measured between generated samples and the validation set. Change relative to the original SD generations is shown. DOODL achieves a more similar set of images in all instances, while the baseline fails in all. }
    \label{tab:fgvc_fid_results}
\end{table}

\subsection{Vocabulary Expansion}
Fine-grained visual classification (FGVC) models aim to classify datasets with subtle variations between classes, such as the Caltech-UCSD Birds (CUB)~\cite{cub}, Stanford Dogs (Dogs)~\cite{dogs},  and FGVC-Aircraft (FGVC-A)~\cite{aircraft} datasets.  
We seek to expand and refine the vocabulary of a pretrained DDM using DOODL such that it can generate a specific class learnt by an FGVC classifier.
For classifier guidance, we use binary cross-entropy (BCE) loss from supervised models trained on FGVC datasets. 
Given a model $m$ trained on a dataset with classes $\{class_i\}_{i=1}^{n_c}$ we generate instances of $class_j$ via loss $c(x^{(i)}_0)=\lambda BCE(m(x^{(i)}_0), j)$, $(\lambda_{\tiny{DOODL}}, \lambda_{Baseline})=(0.05,5)$. 

We evaluate DOODL's ability to expand Stable Diffusion's vocabulary using FGVC classifiers  trained, respectively, on CUB, Dogs, and FGVC-A using WS-DAN~\cite{wsdan}.
The majority of concepts present in these datasets are rare or non-existent in the training data and as such, the pretrained DDM cannot generate accurate images for them (\cref{fig:fgvc_qualitative}-second row). 
We measure the FID~\cite{fid} between a set of generated images (4 seeds) and the validation set of the FGVC dataset being studied. Despite overlap between ImageNet and the targeted datasets, FID has historically still been considered a useful generative evaluation metric~\cite{infogan}. In \cref{tab:fgvc_fid_results} we see that DOODL reduces FID compared to original Stable Diffusion generation on all datasets (average decrease of $\sim3\%$), while classifier guidance does not improve the original generation FID on {any}. 
This indicates that DOODL is able to better incorporate the gradient signal from the classifiers as compared to the one-step approximation. Indeed, example images (\cref{fig:fgvc_qualitative}) show that DOODL can produce images with requisite fine-grained features that identify the object category in question (e.g., a bird species or a dog breed), such as colors, textures, and shapes. DOODL obtains the least relative improvement on Aircrafts, perhaps due to the class differences being largely subtle structural changes with very few textural cues that can act as guiding signals for the optimization process.

\begin{figure}
    \centering
    \includegraphics[width=\columnwidth]{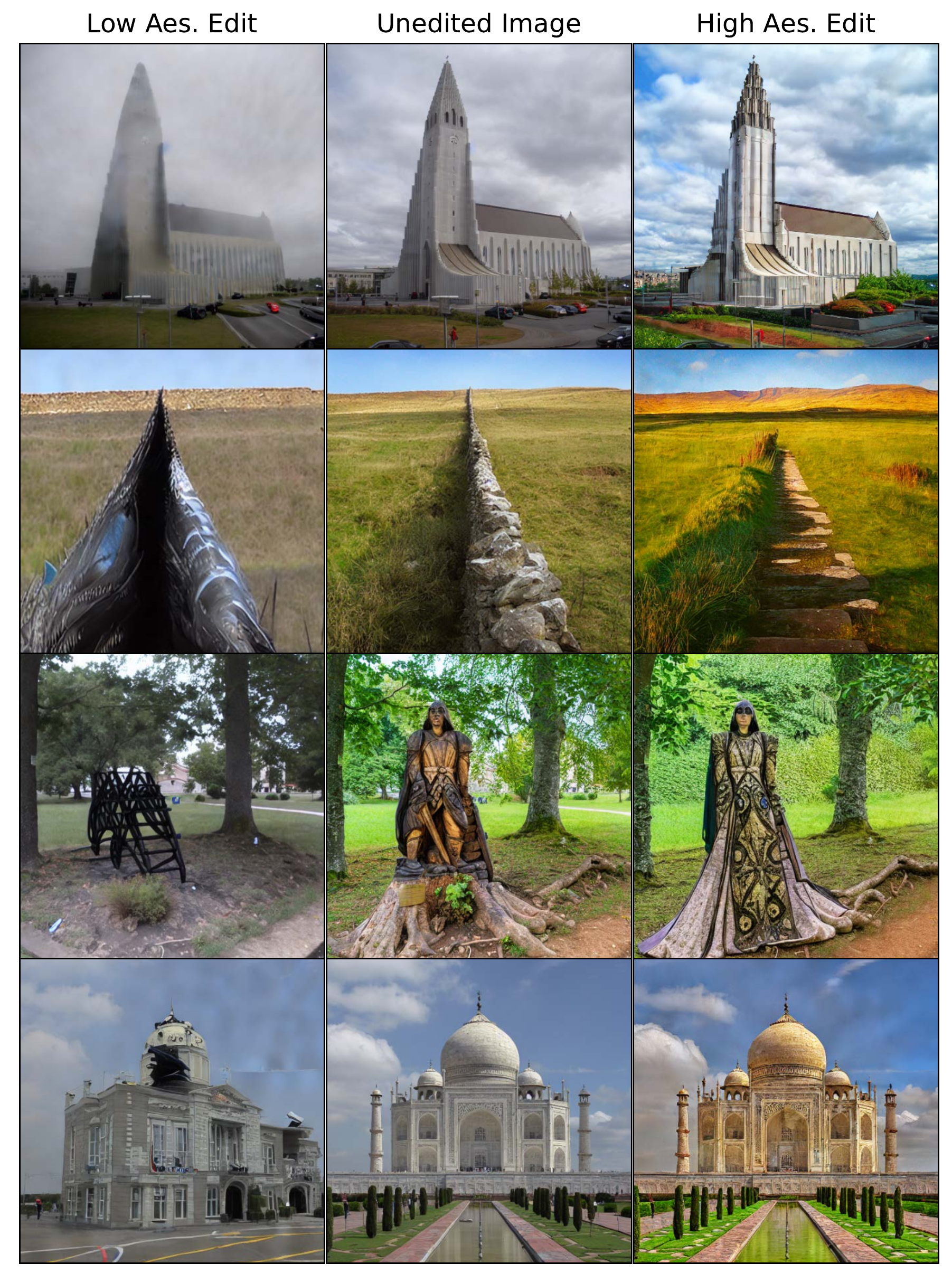}
    \caption{We use EDICT to invert a real-world image to latents which are then optimized using DOODL, targeting an ``aesthetic score'' of either 1 (top row) or 10 (bottom).}
    \label{fig:aes_edits}
\end{figure}

\subsection{Aesthetic Improvement}\label{sec:aes}
Image generation research has emphasized the ability to generate aesthetically pleasing images. 
Notably, the publicly available Stable Diffusion model is trained on the LAION-5B dataset pre-filtered using an "Aesthetics Predictor" model. 
The "Aesthetics Predictor" model $a$ is a linear head trained on top of CLIP visual embeddings to predict a scalar value in the range of [1,10], indicating perceived aesthetic quality (trained on ~\cite{sac}). 
This can easily be incorporated into DOODL's framework with the cost function $c_{aes}(x^{(i)}_0)=\lambda \cdot |a(x^{(i)}_0) - A|,\ \lambda_{DOODL}=1$ where $A=10$ for aesthetic maximization and $1$ for minimization. %

\begin{figure}
    \centering
    \includegraphics[width=\linewidth]{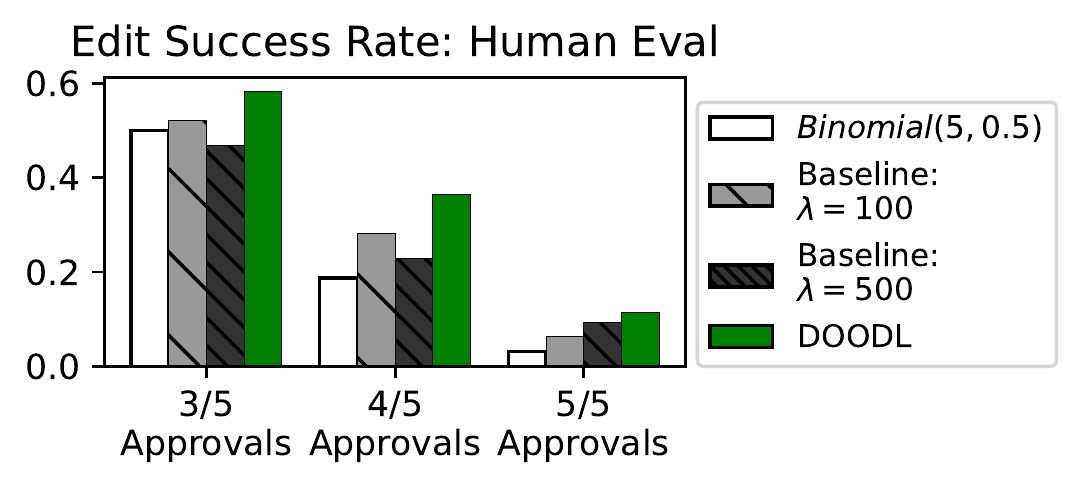}
    \caption{
    Quantitative evaluation of aesthetic quality editing.
    5 labelers are asked whether the edited image both retains content and improves perceptual appeal. 
    Displayed are the success rates for DOODL and the baseline method (two guidance scale settings) for 3 increasingly stringent approval thresholds.
    Accounting for labeling noise, we plot the ``success rate'' of a $Binomial(5, 0.5)$ distribution.
    Evaluation is done on 96 random images from COCO~\cite{lin2014microsoft}.
}
    \label{fig:aes_edit_quant}
\end{figure}

While we find improvement signal for incorporating $c_{aes}$ into text-to-image generations, we place the results in the Supplementary and instead focus on \textit{editing} an existing image to \textit{increase} its perceptual appeal, a novel task with a practical application for users who may wish to improve the aesthetic quality of their photographs. 
We employ EDICT to unconditionally invert an existing image to a latent pair $(x_T,y_T)$ and then optimize the latents w.r.t. $c_{aes}$.%
Qualitative results of incorporating this cost function are shown in \cref{fig:aes_edits} where in-the-wild images are edited to have lower or higher aesthetic quality. We quantify DOODL's editing ability using human evaluation in \cref{fig:aes_edit_quant}, seeing that DOODL produces satisfactory edits significantly more often than the baseline classifier guidance.
For the baselines in the latter, we find that inverting the images to $t=T$ results in generation content largely decorrelated from the original image, and invert to $t=\frac{T}{2}$ which we find to produce satisfactory edits.

\begin{figure}
    \centering
    \includegraphics[width=\linewidth]{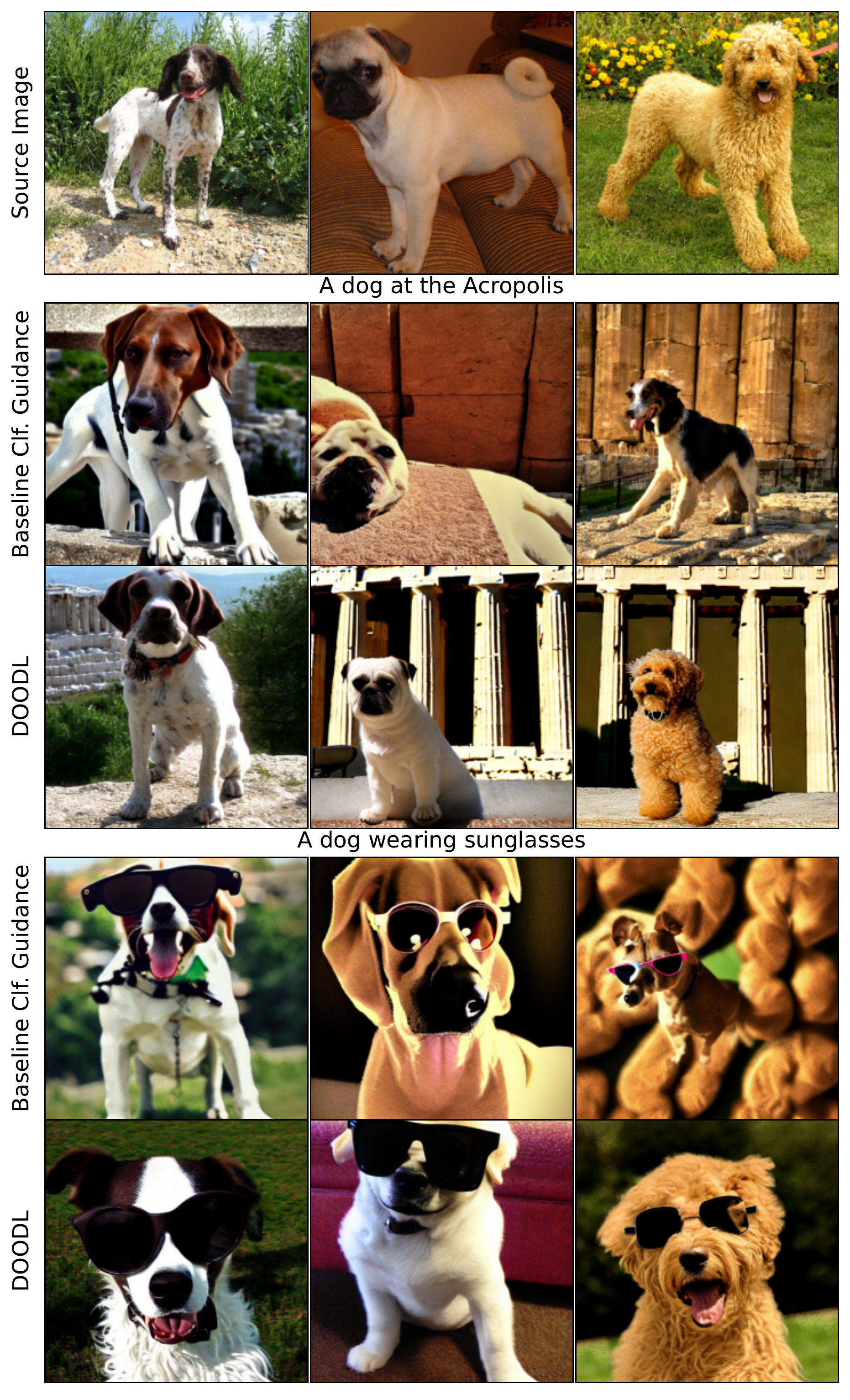}
    \caption{Personalization generations from a single seed. 
    Baseline Clf. Guidance does not depict the pictured dog without destabilizing. DOODL produces realistic images containing both the desired text and image conditioning. }
    \label{fig:visual_qualitative}
\end{figure}

\subsection{Visual Personalization Guidance}\label{sec:visual_method}
Visual personalization, making a diffusion model generate images that contain a highly specific entity or concept based off of a limited exemplar set of images, is an area that has garnered much research attention.
Most methods, such as Dreambooth~\cite{dreambooth} or Textual Inversion~\cite{textualinversion} learn or finetune components such as text embeddings or subsets of the diffusion model parameters in order to generate visually personalized images.
While highly effective, the training cost and subsequent model storage are drawbacks.
We propose a novel on-the-fly personalization paradigm using DOODL to employ a pretrained recognition model in leiu of \textit{any} additional generative model training or tuning.

We utilize the distance from \cref{eq:spherical_dist}  taken between the image embedding of a conditioning image, $CLIP_{Image}(C_{Image})$, and that of the current generation $CLIP_{Image}(x^{(i)}_0)$
We find ensembling ViT/B-32, ViT/L-14, and ViT/g-14 CLIP models with loss weights of 0.5, 0.25, 0.25 respectively achieves good performance.  Input text conditioning to the diffusion model is as standard.
A similar regime of $\lambda_{DOODL}$ performs here as in \cref{sec:text_guidance_method}, $\lambda_{DOODL}=3$ for displayed results. This setting is unusually difficult for DOODL, requiring more ($5\times$)  optimization iterations than other reported experiments (additionally, we find a much higher $\lambda_{Baseline}$ needed).

We visualize generated samples in \cref{fig:visual_qualitative}, placing dogs pictured in the top row ($C_{Image}$) into various contexts. Prompts generically refer to ``a dog", with all identifying information coming from $C_{Image}$. DOODL maintains faithfulness to both the text prompt and visual conditioning; a first for non-learned methods to our knowledge.

We perform a human evaluation, sampling four target dog images from Imagenet~\cite{imagenet} and using four prompt templates (\textit{A dog \{at the Acropolis/swimming/in a bucket/wearing sunglasses\}}) across 6 random seeds, resulting in 96 generated images per method. For each image, three labelers are asked whether the dog in the generated image appears to be the same dog as the original with the context matching the caption. A majority of labelers classify the original generation and baseline as successful just $2\%$ and $1\%$ of the time respectively as opposed to $14\%$ for DOODL, $7\times$ more often than either baseline. Further analysis is given in the Supplementary.

\section{Conclusion \& Future Work}
\NNnote{to revisit}
In this work, we demonstrated that Direct Optimization Of Diffusion Latents (DOODL) offers an exciting new way to incorporate the knowledge of pretrained recognition networks in the generative process of diffusion models. We expect future work to both expand the types of guidances incorporated as well as sophisticate and accelerate the optimization process of DOODL to incorporate the process into applications that require higher compute efficiency.

{\small
\bibliographystyle{ieee_fullname}
\bibliography{main}
}

\clearpage

\crefname{section}{Sec.}{Secs.}
\Crefname{section}{Section}{Sections}
\Crefname{table}{Table}{Tables}
\crefname{table}{Tab.}{Tabs.}
\crefname{figure}{Fig. S}{ Fig. S}

\crefformat{figure}{Figure S#2#1#3}
\crefformat{table}{Table S#2#1#3}
\renewcommand\figurename{Figure S}
\renewcommand\thefigure{\unskip\arabic{figure}}

\renewcommand\tablename{Table S}
\renewcommand\thetable{\unskip\arabic{table}}

\appendix

This supplementary materials contains
\begin{enumerate}
    \item Results for improving aesthetic appeal in text-to-image generation (\cref{sec:gen_aes})
    \item Further analysis of personalization results (\cref{sec:personalization})
    \item Details of the DOODL optimization process (\cref{sec:details})
    \item Additional qualitative samples (\cref{sec:samples})
\end{enumerate}

\section{Text-to-Image Generative Aesthetics Results}\label{sec:gen_aes}

\begin{figure*}[t]
    \centering
    \includegraphics[width=\linewidth]{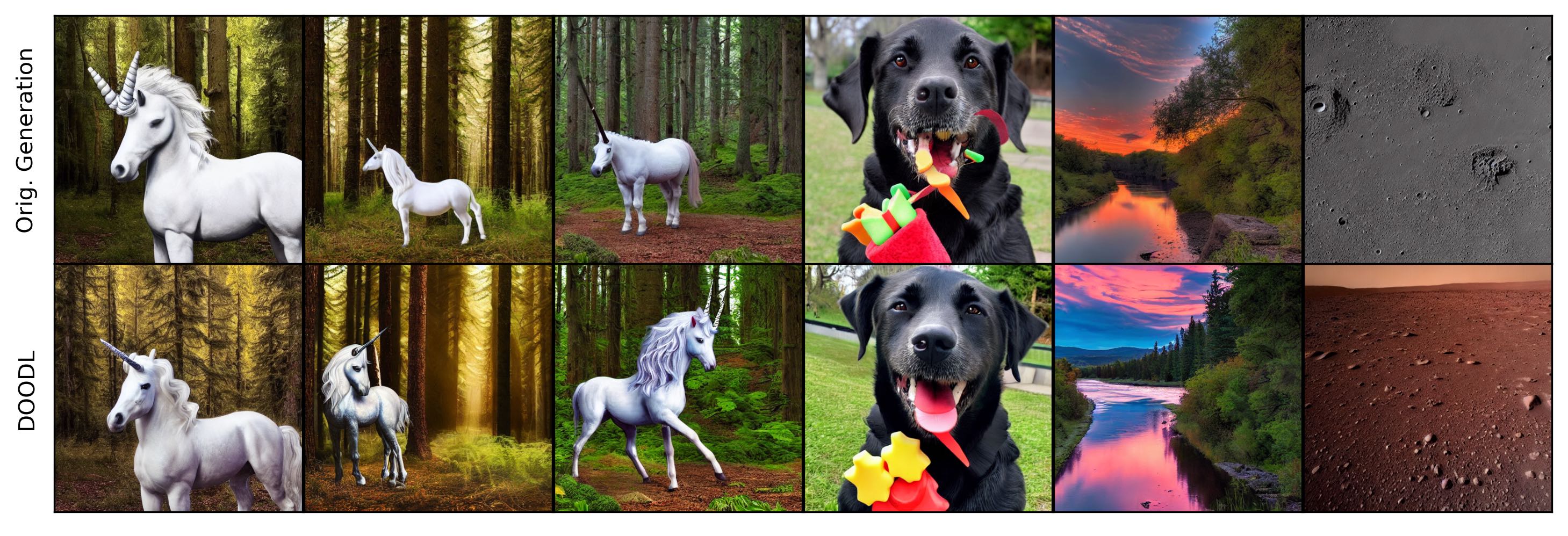}
    \caption{Qualitative Aesthetic Results.
    Prompts: \textit{A unicorn in forest}($\times 3$), \textit{A dog with a chewtoy}, \textit{A river at sunrise}, \textit{A dramatic photo from the surface of Mars}.}
    \label{fig:aesthetic_qualitative}
\end{figure*}

We investigate using DOODL to guide text-to-image generation using the aesthetic scoring model. The standard algorithm is used with $\lambda_{DOODL}=0.1$. 
Qualitative results are shown in \cref{fig:aesthetic_qualitative}.

We additionally perform human evaluation using the same comparison method as in the main paper, asking \textit{Please select the image which you think would be preferred visually by the majority of people}. We sample 8 prompts over 16 seeds:
\begin{itemize}
    \item A surfer catching a wave
    \item A unicorn in forest
    \item A stained glass window
    \item Yosemite valley
    \item A dramatic photo from the surface of mars
    \item A cottage in the countryside
    \item A river at sunrise
    \item A dog with a chewtoy
\end{itemize}

\begin{figure}[t]
    \centering
 \includegraphics[width=\linewidth]{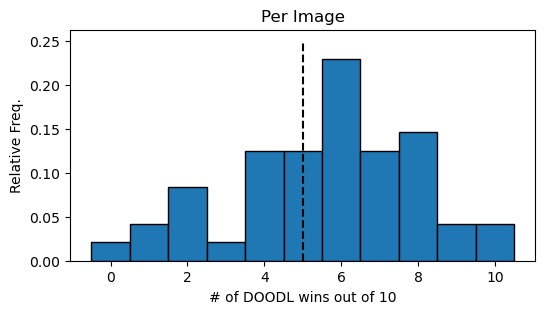}
    \caption{Human Aesthetic Results. 
    10 independent labelers are instructed \textit{Please select the image which you think would be preferred visually by the majority of people}. 
    A ``No preference'' response is given which we count as half a response in each direction.
    The per-image DOODL wins out of 10 is shown.
    The win-draw-loss rates of DOODL are $0.58-0.11-0.31$.}
    \label{fig:aesthetic_human}
\end{figure}

Quantitative results against the original generation are shown in \cref{fig:aesthetic_human}. We find that DOODL generations are consistently rated as having higher aesthetic appeal vs. the original generations with a win-draw-loss rate of $0.58-0.11-0.31$.
We perform the same experiment against baseline classifier guidance with $\lambda_{Baseline}=10$. We find the signal to be less consistent, with a win-draw-loss rate for DOODL of $0.48-0.07-0.45$ across the 128 comparisions. We hypothesize that this is possibly attributable to the value that the aesthetic model places on vivid colors and contrast; lower-level features that don't neccessarily require the precision of DOODL's approach vs. more approximate control. DOODL also is more prone to warp or deform content than the baseline guidance in performing the model-based optimization, which is typically visually unappealing. Further stabilizing DOODL with respect to the latter point would serve to broaden the above performance gap.

\section{Further Analysis of Personalization Results}\label{sec:personalization}

In the main text, we presented how often the majority (2/3) labelers labeled the dog as appearing to be the original dog in the desired context.
The responses labelers chose from were:

\begin{enumerate}
    \item The second image does not match the prompt, or is highly unrealistic
    \item The dog in the second image does not look like the original dog
    \item The dog in the second image looks similar to the original dog but there are significant differences
    \item The dog in the second image appears to be the original dog
\end{enumerate}

We present several different views of the data, all confirming that DOODL achieves far superior performance to the baselines and opens a door to a new family of guidance-based personalization methods.

\paragraph{Aggregate Statistics}

4.5\% of the original generations were labeled (4), as opposed to 5.6\% for the baseline and 19.4\% for DOODL.

\paragraph{Unanimous Agreement}

Due to the challenging problem and noisiness in the labeling process, very few images were unanimously classified as (4). 3.1\% (3/96) were for DOODL and 0 for the other two methods.

\paragraph{Lowering Similarity Bar}
We visualize the success rates if allowing responses (3) or (4) (so the dog must look similar but differences are allowed) in \cref{tab:personalization}.

\begin{table*}[]
    \centering
    \begin{tabular}{ c | c c c}
    Method & Original & Baseline Clf. Guidance & DOODL \\
    \hline
         Aggregate Statistics & 12.2\% & 28.5\% & 52.1\% \\
         Majority Agreement &  6.2\% & 25\% & 52.1\% \\
         Unanimous Agreement & 0\% & 8.3\% & 25\% 
    \end{tabular}
    \caption{Success rates for personalization with lowering criteria to \textit{The dog in the second image looks similar to the original dog but there are significant differences}}
    \label{tab:personalization}
\end{table*}

\section{Details on Multicrop in DOODL Optimization}\label{sec:details}

Here we precisely describe the multicrop augmentation used in the DOODL optimization process. We employ the same MakeCutouts class as used in the diffusers library\cite{diffuserslibrary}, with code in \cref{fig:code}.
We use \texttt{cut\_power}=0.3.
The size of the square crop is sampled by generation a random value $r \sim U(0,1)$ and crop size $ModelInputSize + (OriginalImageSize - ModelInputSize) r^{CutPower}$. A crop of this size is then uniformly selected from the image.

\begin{figure*}
\begin{verbatim}
class MakeCutouts(nn.Module):
    def __init__(self, cut_size, cut_power=1.0):
        super().__init__()

        self.cut_size = cut_size
        self.cut_power = cut_power

    def forward(self, pixel_values, num_cutouts):
        sideY, sideX = pixel_values.shape[2:4]
        max_size = min(sideX, sideY)
        min_size = min(sideX, sideY, self.cut_size)
        cutouts = []
        for _ in range(num_cutouts):
            size = int(torch.rand([]) ** self.cut_power * (max_size - min_size) + min_size)
            offsetx = torch.randint(0, sideX - size + 1, ())
            offsety = torch.randint(0, sideY - size + 1, ())
            cutout = pixel_values[:, :, offsety : offsety + size, offsetx : offsetx + size]
            cutouts.append(F.adaptive_avg_pool2d(cutout, self.cut_size))
        return torch.cat(cutouts)
\end{verbatim}
\caption{Multicrop python code from \cite{diffuserslibrary} examples.}
\label{fig:code}
\end{figure*}

\section{Additional Qualitative Results}\label{sec:samples}

Additional  qualitative results are given in the below listed figures. 

\begin{figure*}
    \centering
\includegraphics[width=0.9\linewidth]{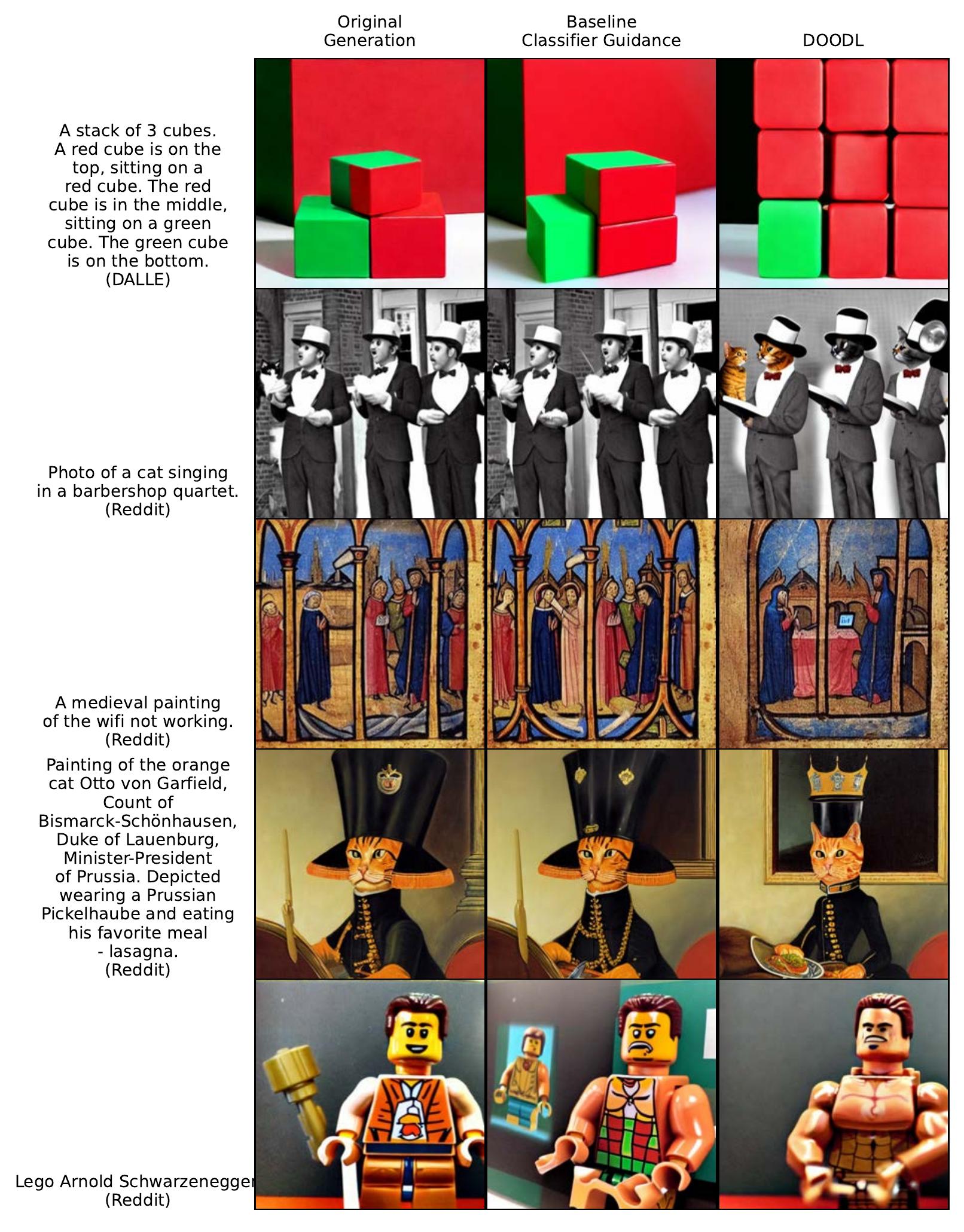}
    \caption{Additional Drawbench results}
    \label{fig:drawbench}
\end{figure*}

\begin{figure*}
    \centering
\includegraphics[width=\linewidth]{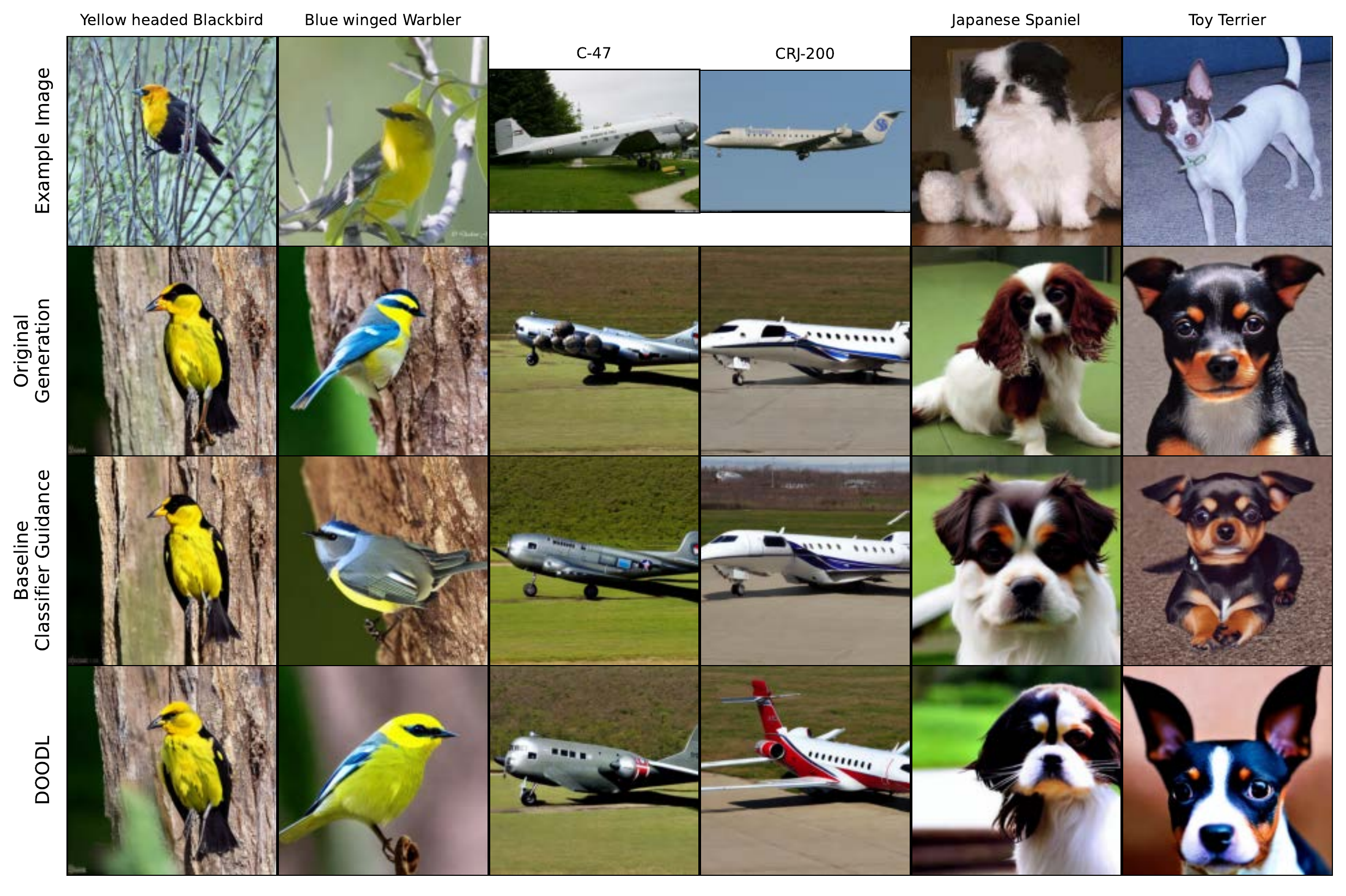}
    \caption{Additional FGVC results}
    \label{fig:fgvc}
\end{figure*}

\begin{figure*}
    \centering
    \includegraphics{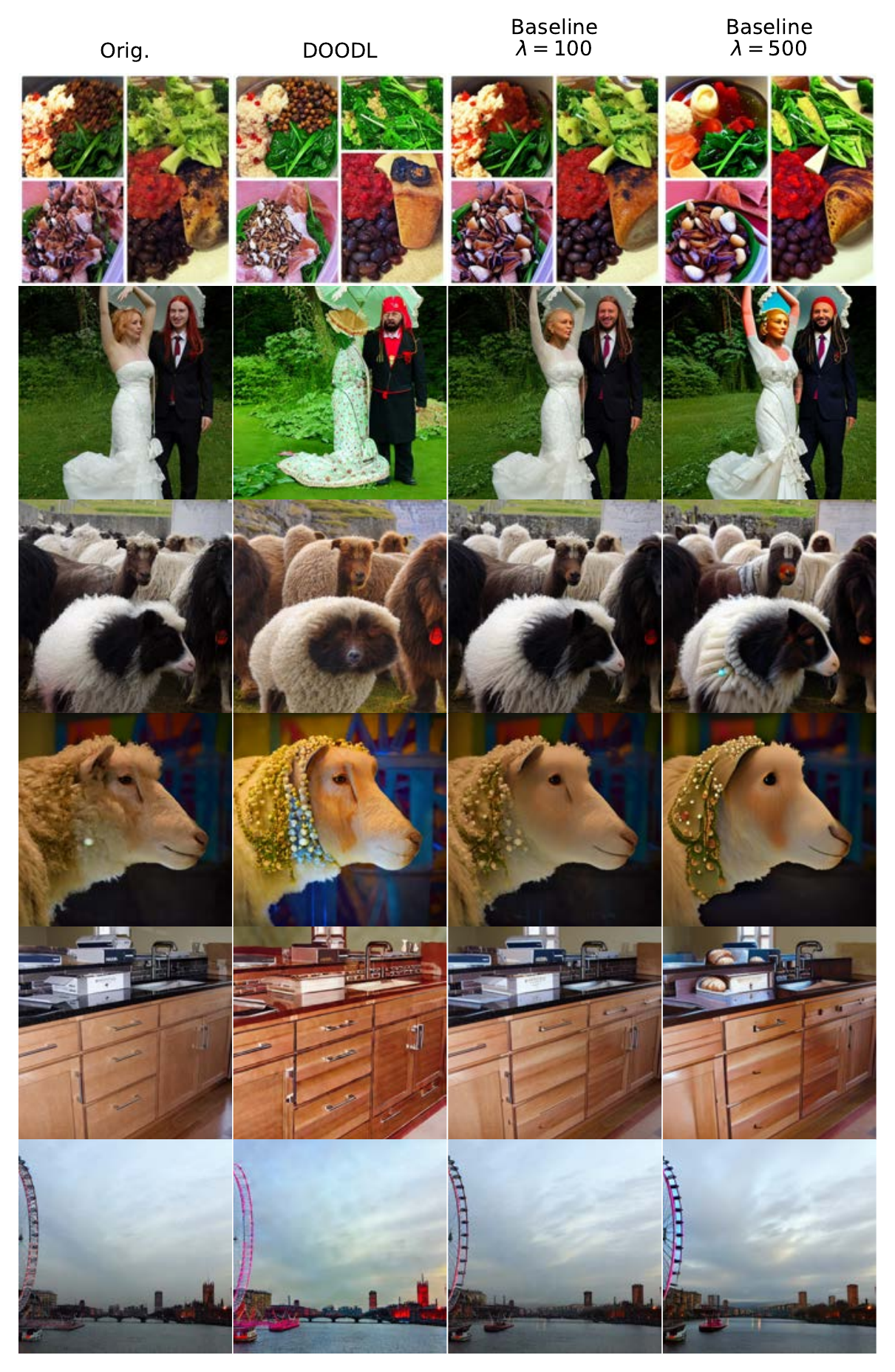}
    \caption{Additional random COCO aesthetic editing results (1). No caption information is used}
    \label{fig:supp_aes_edit_1}
\end{figure*}

\begin{figure*}
    \centering
    \includegraphics{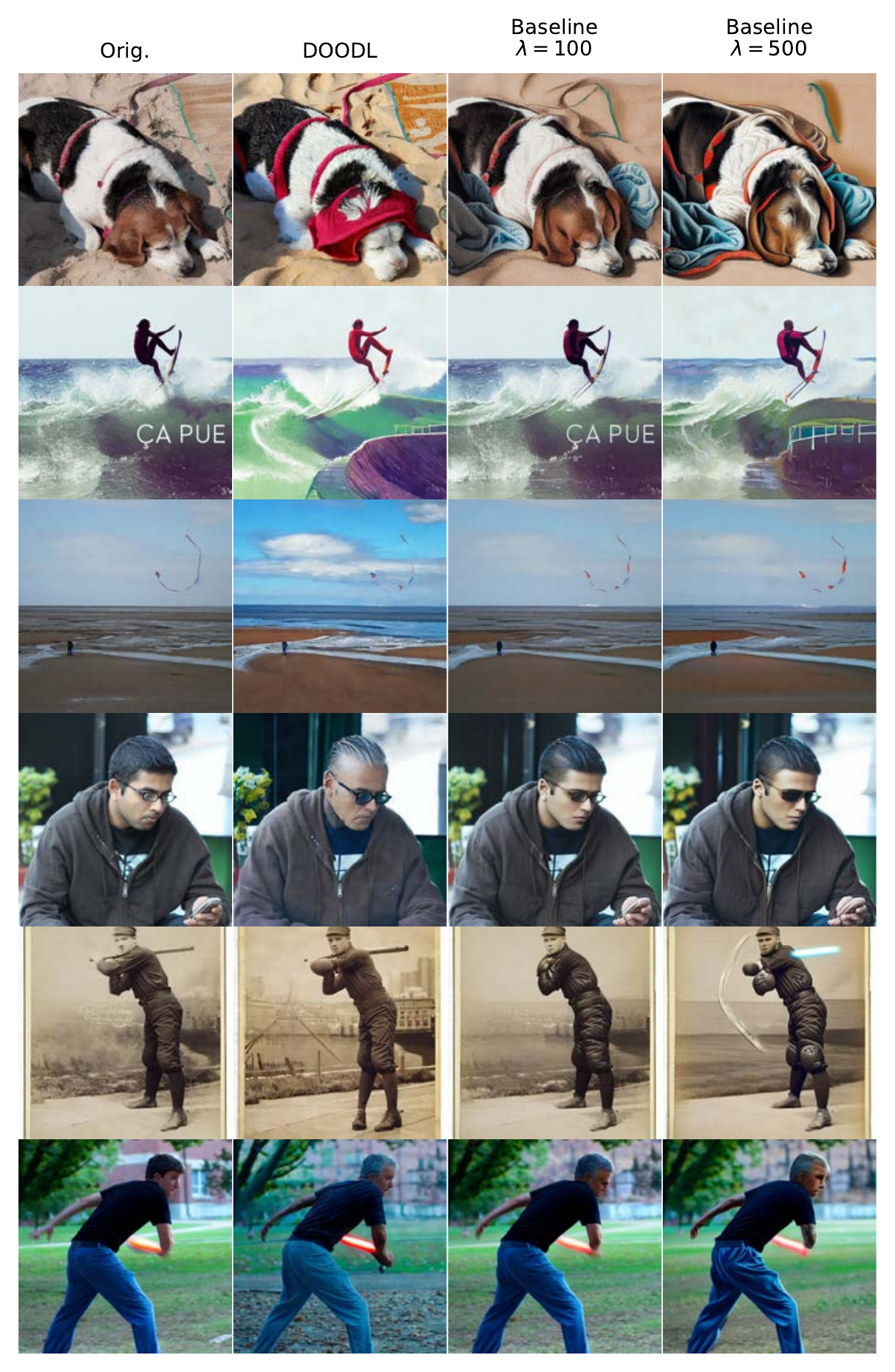}
    \caption{Additional random COCO aesthetic editing results (2). No caption information is used}
    \label{fig:supp_aes_edit_2}
\end{figure*}

\begin{figure*}
    \centering
    \includegraphics{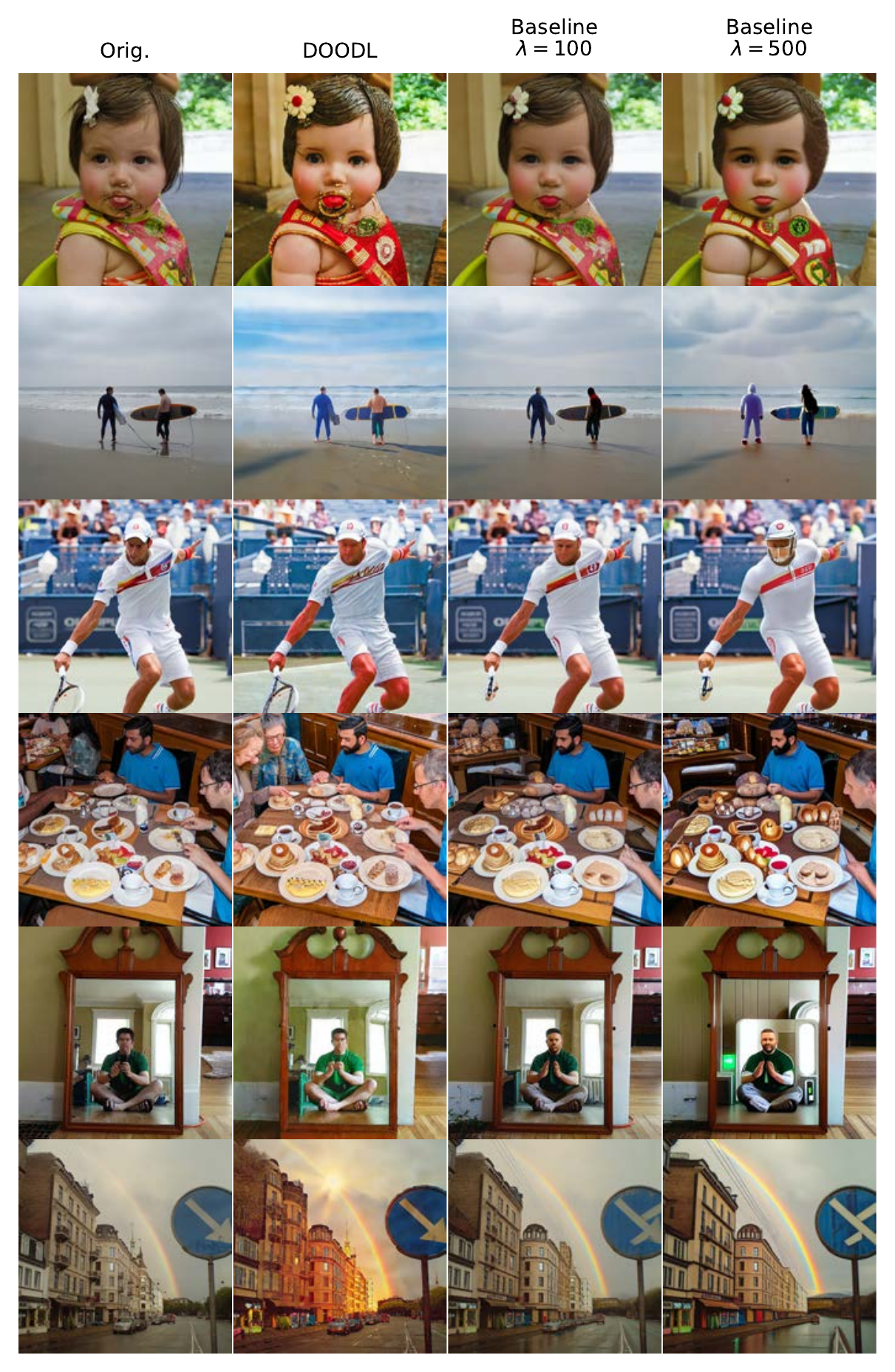}
    \caption{Additional random COCO aesthetic editing results (3). No caption information is used}
    \label{fig:supp_aes_edit_3}
\end{figure*}

\begin{figure*}
    \centering
    \includegraphics{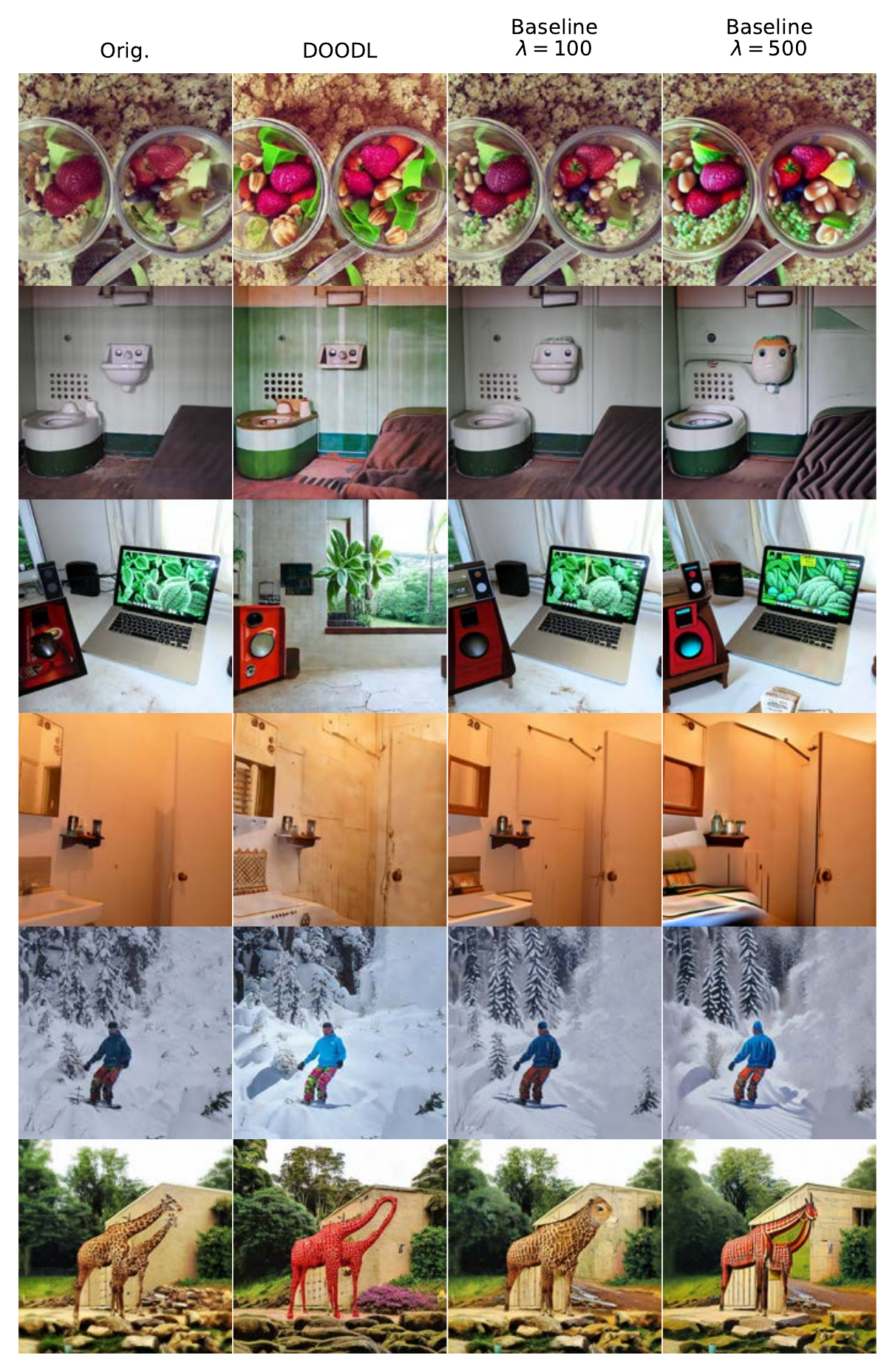}
    \caption{Additional random COCO aesthetic editing results (4). No caption information is used}
    \label{fig:supp_aes_edit_4}
\end{figure*}

\begin{figure*}
    \centering
    \includegraphics[width=\linewidth]{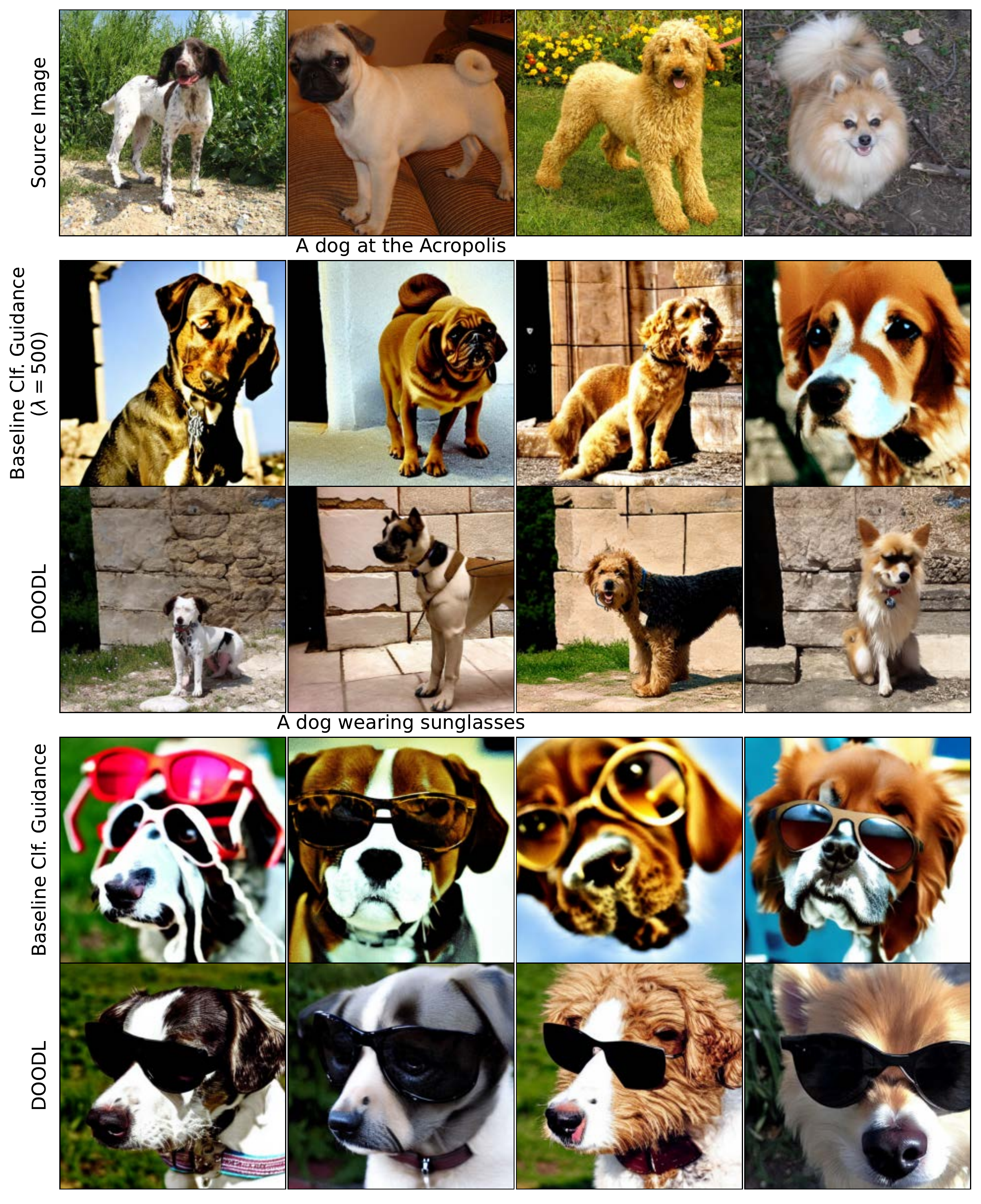}
    \caption{Additional personalization results}
    \label{fig:personalization_0}
\end{figure*}

\begin{figure*}
    \centering
    \includegraphics[width=\linewidth]{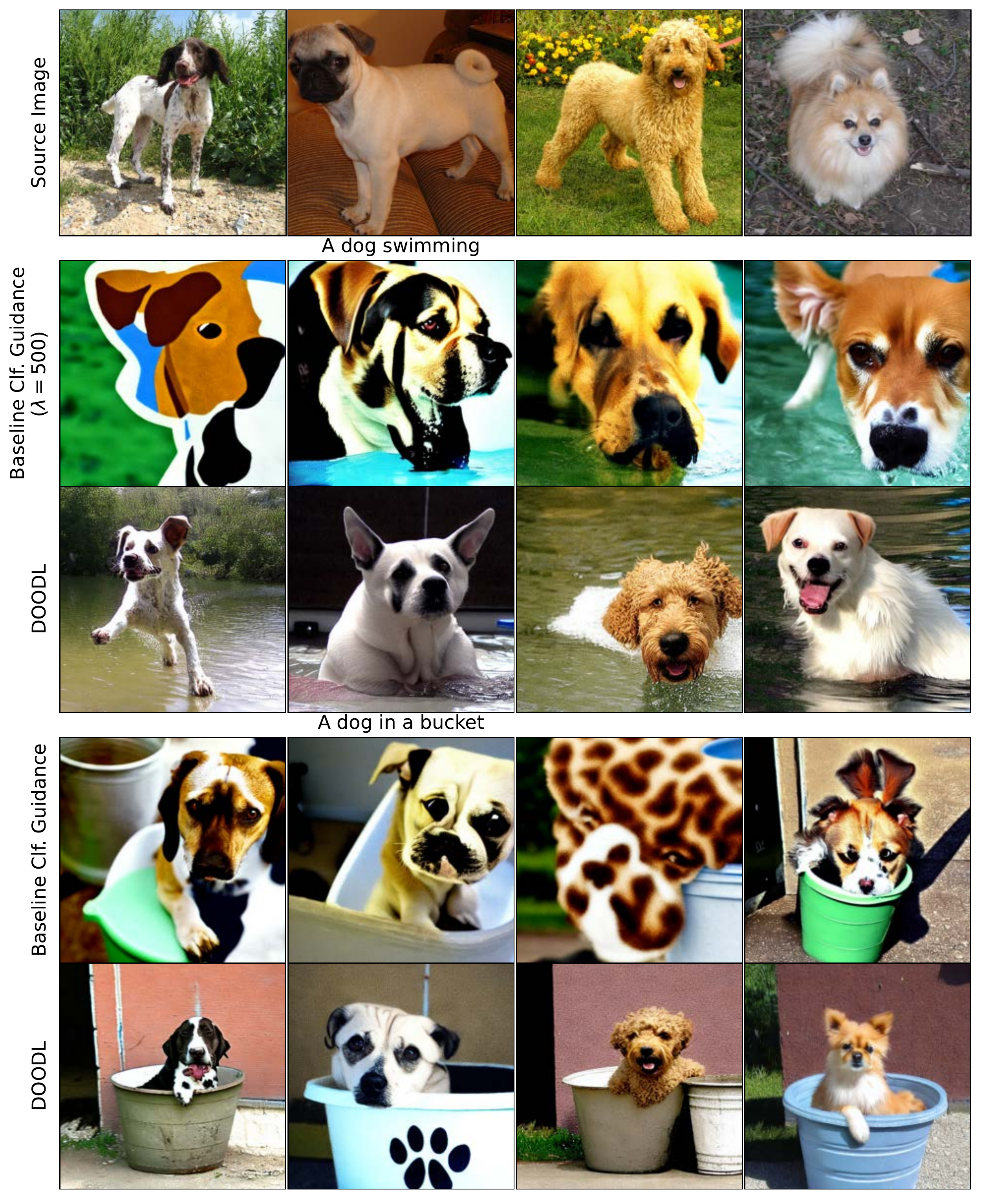}
    \caption{Additional personalization results}
    \label{fig:personalization_1}
\end{figure*}

\begin{figure*}
    \centering
    \includegraphics[width=\linewidth]{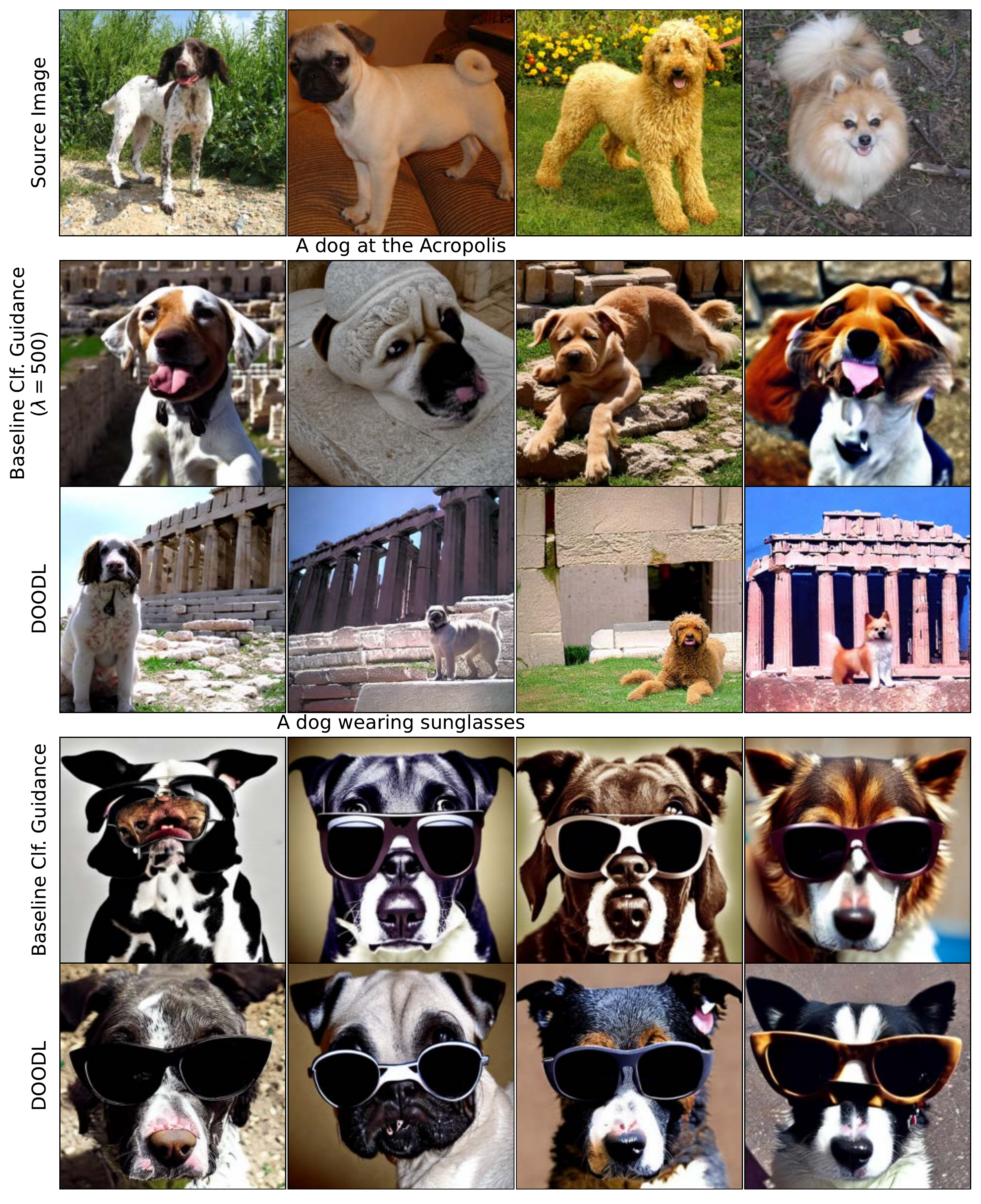}
    \caption{Additional personalization results}
    \label{fig:personalization_2}
\end{figure*}

\begin{figure*}
    \centering
    \includegraphics[width=\linewidth]{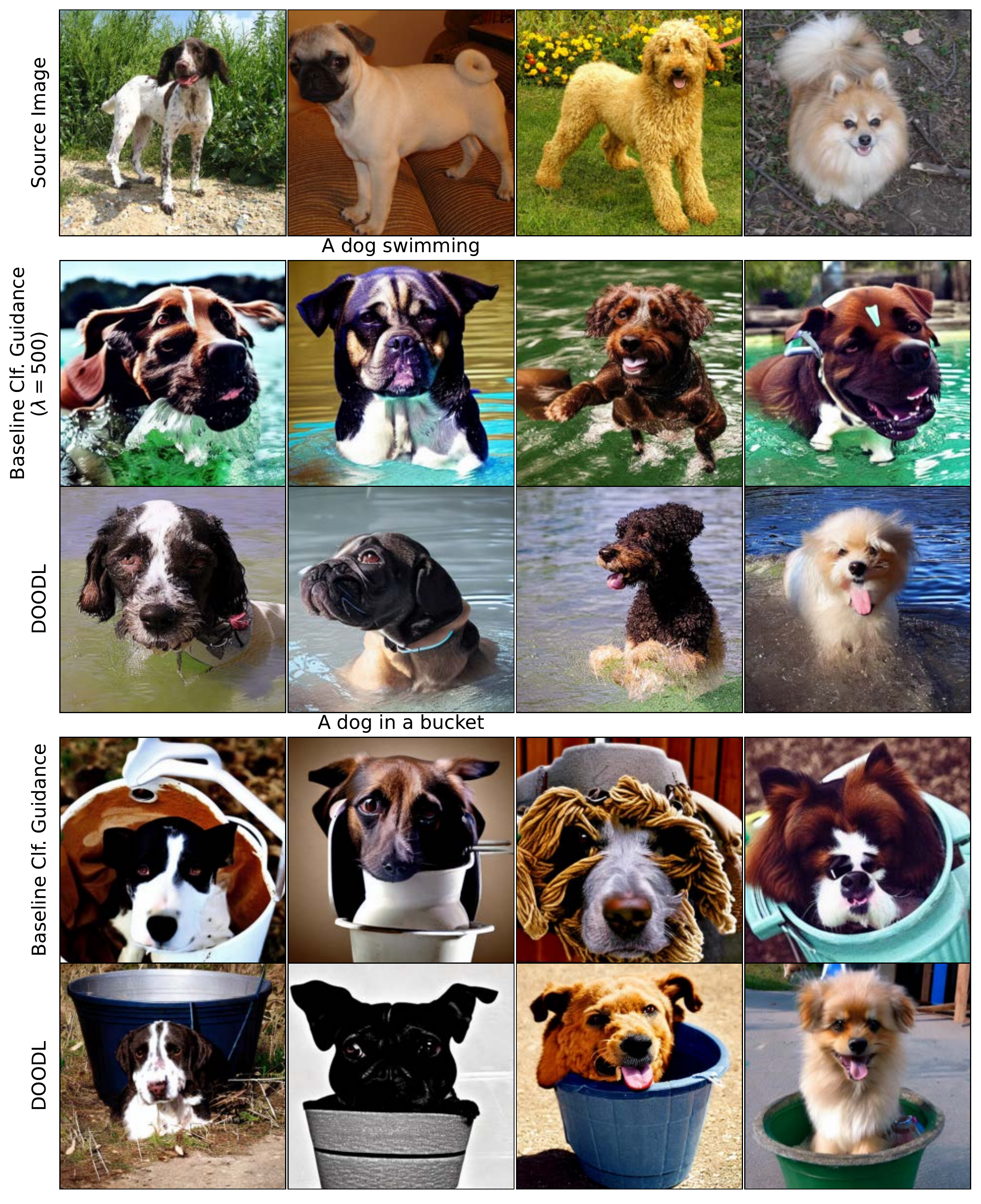}
    \caption{Additional personalization results}
    \label{fig:personalization_3}
\end{figure*}

\begin{figure*}
    \centering
    \includegraphics[width=\linewidth]{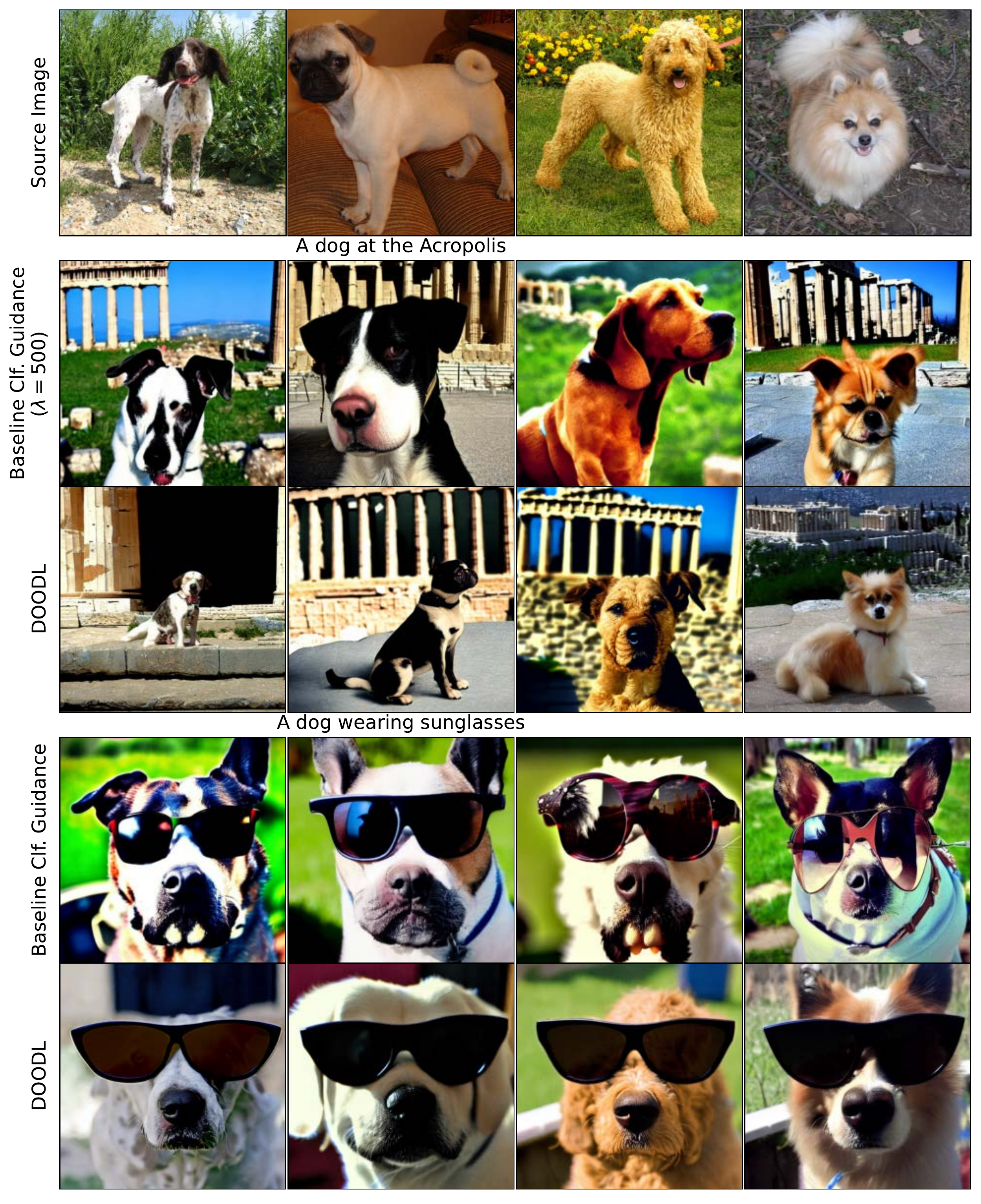}
    \caption{Additional personalization results}
    \label{fig:personalization_4}
\end{figure*}

\begin{figure*}
    \centering
    \includegraphics[width=\linewidth]{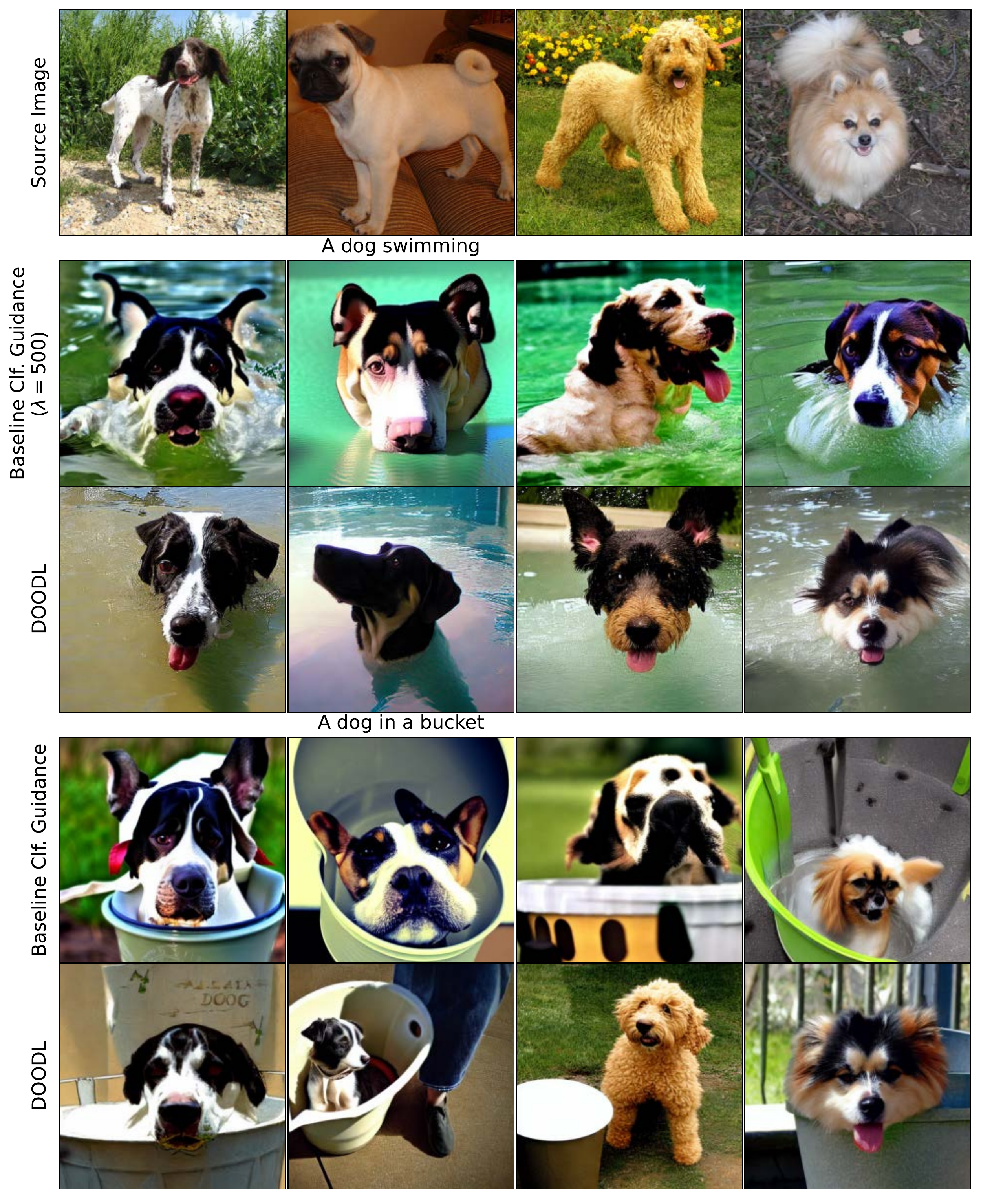}
    \caption{Additional personalization results}
    \label{fig:personalization_5}
\end{figure*}

\begin{figure*}
    \centering
    \includegraphics[width=\linewidth]{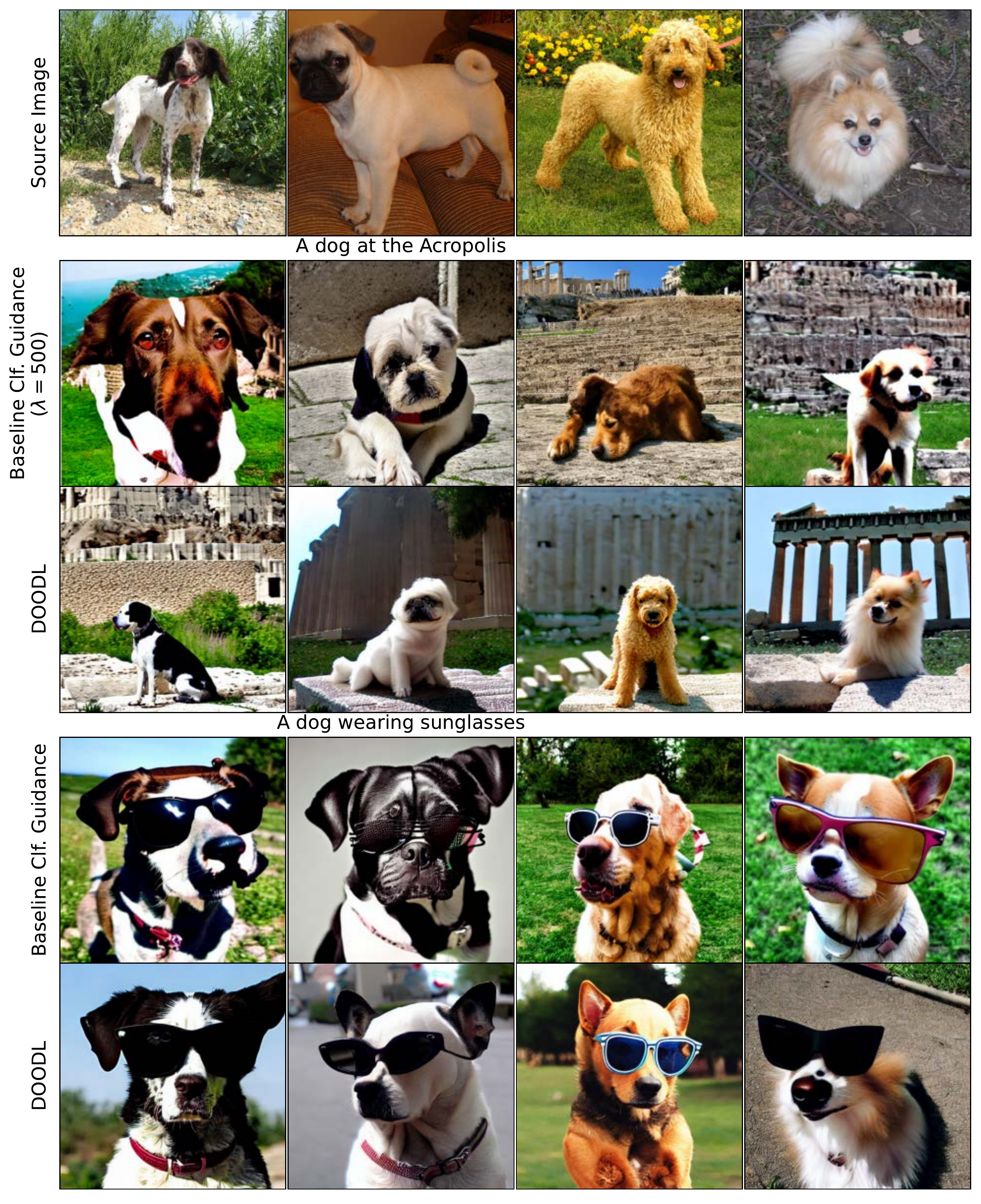}
    \caption{Additional personalization results}
    \label{fig:personalization_6}
\end{figure*}

\begin{figure*}
    \centering
    \includegraphics[width=\linewidth]{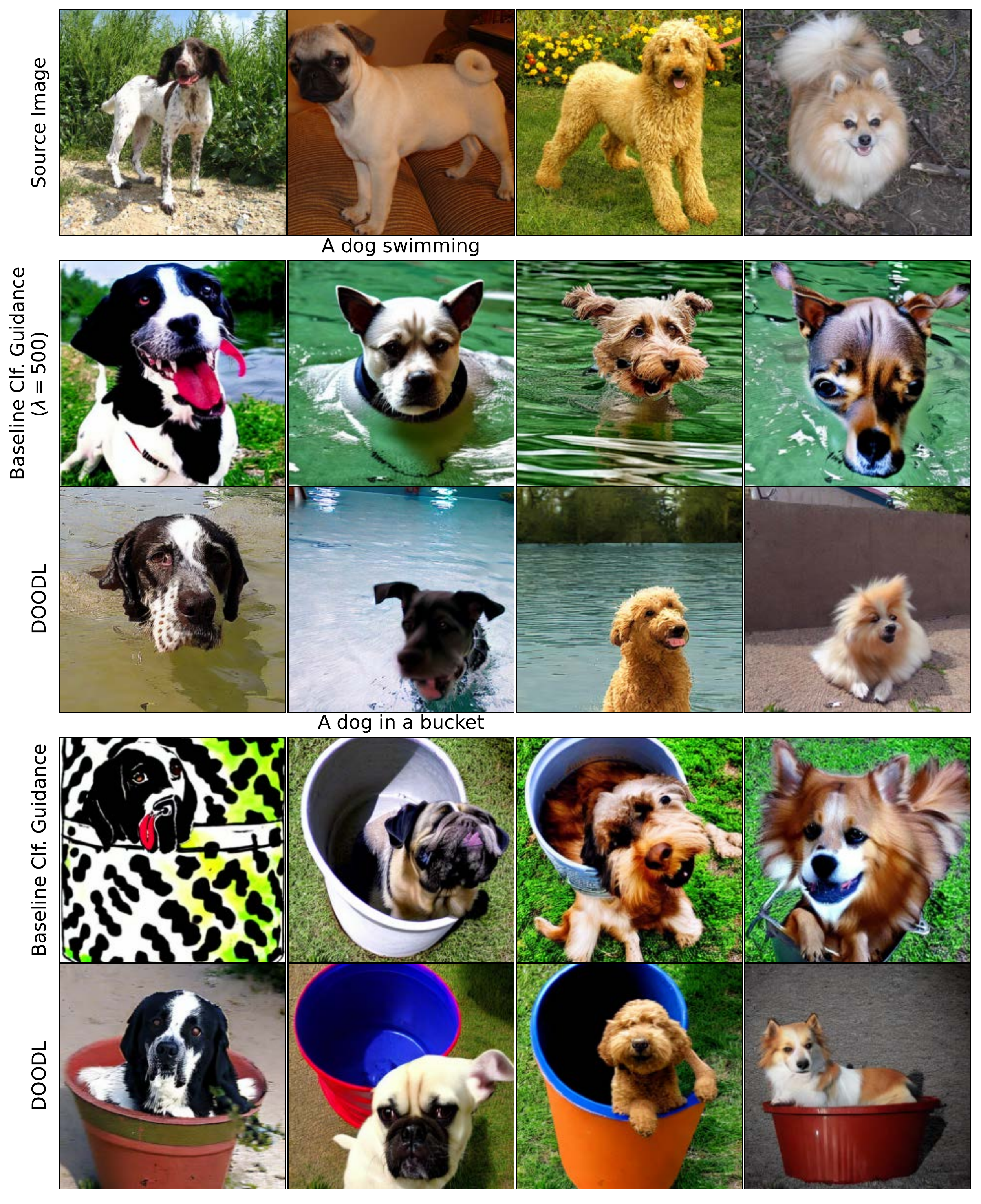}
    \caption{Additional personalization results}
    \label{fig:personalization_7}
\end{figure*}

\begin{itemize}
    \item \cref{fig:drawbench} for Drawbench generations %
    \item \cref{fig:fgvc} for additional FGVC results
    \item \cref{fig:supp_aes_edit_1,fig:supp_aes_edit_2,fig:supp_aes_edit_3,fig:supp_aes_edit_4} for randomly chosen aesthetic editing on COCO from the result set used in human evaluation for Section 5.3.
    \item \cref{fig:personalization_0,fig:personalization_1,fig:personalization_2,fig:personalization_3,fig:personalization_4,fig:personalization_5,fig:personalization_6,fig:personalization_7} for further personalization results across source images and target captions with random seeds.
\end{itemize}



\end{document}